%% file: iclr2027_conference.tex
\documentclass{article} %
\usepackage{iclr2027_conference,times}

\input{math_commands.tex}

\usepackage{hyperref}
\usepackage{url}
\usepackage{booktabs}
\usepackage{array}
\usepackage{colortbl}
\usepackage{xfp}
\usepackage{graphicx}
\usepackage{float}
\usepackage{placeins}
\usepackage{algorithm}
\usepackage{algpseudocode}
\usepackage{listings}
\usepackage{mycommands}
\usepackage{enumitem}
\usepackage{capt-of}
\usepackage{wrapfig}
\usepackage{xcolor}

\usepackage{xspace}
\newcommand{\bench}{\textsc{VVRBench}\xspace}

\title{Verifiable Visual Rewards Transfer from Synthetic Scenes to Natural Prompts}

\iclrfinalcopy
\author{Shuyue Stella Li,\quad Xiaochuang Han,\quad  Yulia Tsvetkov,\quad  Luke Zettlemoyer \\  \vspace{1mm}
University of Washington\\ \vspace{1mm}
\texttt{stelli@cs.washington.edu} \\ \vspace{1mm}
\parbox{0.03\textwidth}{\includegraphics[width=\linewidth]{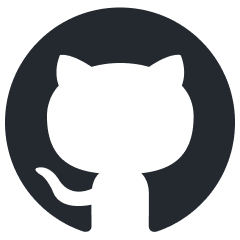}}\hspace{0.5mm}\href{https://github.com/stellalisy/VVRBench}{\texttt{https://github.com/stellalisy/VVRBench}}\\
\hspace{0.7mm}\parbox{0.031\textwidth}{\includegraphics[width=\linewidth]{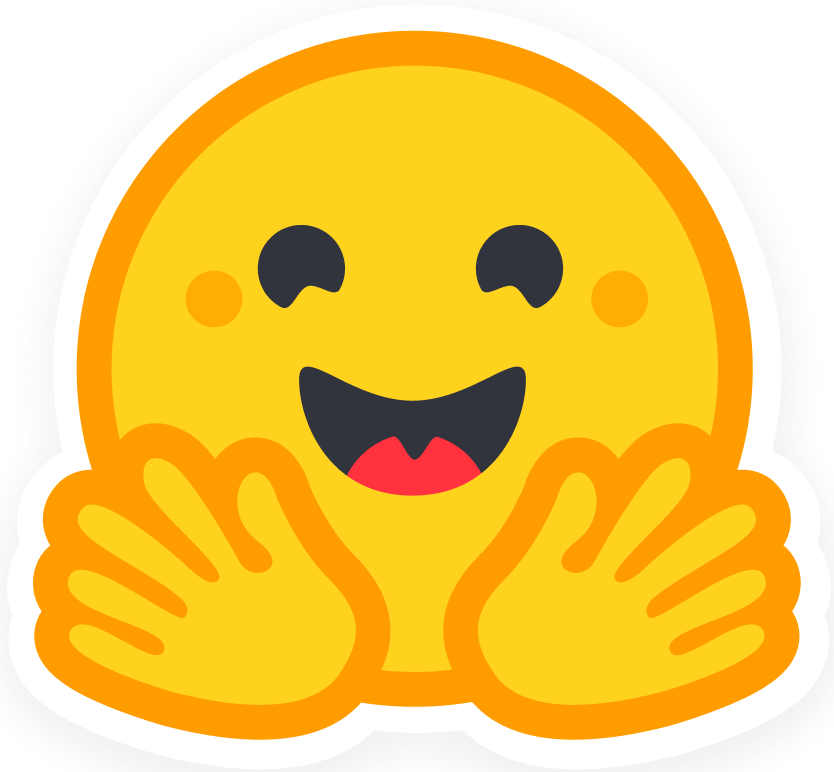}}\hspace{0.5mm}\href{https://huggingface.co/datasets/stellalisy/VVRBench}{\texttt{https://huggingface.co/datasets/stellalisy/VVRBench}}\vspace{-3mm}
}

\definecolor{vvrLow}{RGB}{196,65,57}
\definecolor{vvrHigh}{RGB}{45,115,180}
\newcommand{\vvrbenchscore}[3]{%
  \ifdim#2pt<50pt
    \cellcolor{vvrLow!\fpeval{round(55-0.9*(#2),0)}!white}#3%
  \else
    \cellcolor{vvrHigh!\fpeval{round(0.9*(#2)-35,0)}!white}#3%
  \fi
}
\newtcolorbox[auto counter]{finding}{colback=oc-gray-0,colframe=oc-gray-6,
  boxrule=0.5pt,arc=2pt,left=5pt,right=5pt,top=3pt,bottom=3pt,
  before upper={\textbf{Finding~\thetcbcounter.}\ }}
\definecolor{wrong}{HTML}{b82111}
\definecolor{oc-green-0}{HTML}{EBFBEE}
\definecolor{oc-green-8}{HTML}{2B8A3E}
\newtcolorbox{abstention}[1]{enhanced,colback=white,colframe=oc-gray-6,boxrule=0.5pt,arc=2pt,
  left=4pt,right=4pt,top=3pt,bottom=3pt,fonttitle=\small\bfseries,coltitle=black,colbacktitle=oc-gray-1,
  fontupper=\small,title={#1}}
\newtcolorbox{abstentionresponse}{colback=oc-red-0!35!white,colframe=oc-red-8,boxrule=0.4pt,arc=1pt,
  left=3pt,right=3pt,top=1pt,bottom=1pt,before upper={\textcolor{oc-red-8}{\textbf{Model response.}}\ }}
\newtcolorbox{abstentionwhy}{colback=oc-green-0,colframe=oc-green-8,boxrule=0.4pt,arc=1pt,
  left=3pt,right=3pt,top=1pt,bottom=1pt,before upper={\textcolor{oc-green-8}{\textbf{Why it is wrong.}}\ }}

\newcommand{\cblock}[3]{
  \mbox{
  \protect\hspace{-1.5mm}
  \protect\begin{tikzpicture}
    \protect\node[draw, minimum size=2.5mm, thick, line width=0.5pt, 
          fill={rgb,255:red,#1;green,#2;blue,#3}] () {};
  \protect\end{tikzpicture}%
  }
}

\begin{document}

\maketitle

\input{sec/0_abstract}
\input{sec/1_intro}
\input{sec/2_methods}

\input{sec/3_benchmark}
\input{sec/4_posttraining}
\input{sec/5_related_work}
\input{sec/6_conclusion}

\newpage
\input{sec/7_limitations}
\input{sec/7_statements}

\bibliography{iclr2027_conference}
\bibliographystyle{iclr2027_conference}

\input{app/appendix}

\end{document}

%% file: math_commands.tex
\usepackage{amsmath,amsfonts,bm}

\def\eqref#1{equation~\ref{#1}}

\def\1{\bm{1}}

\DeclareMathAlphabet{\mathsfit}{\encodingdefault}{\sfdefault}{m}{sl}
\SetMathAlphabet{\mathsfit}{bold}{\encodingdefault}{\sfdefault}{bx}{n}

%% file: sec/0_abstract.tex
\begin{abstract}

\vspace{-2mm}

Precise instruction following in image generation, such as satisfying object counts and spatial relations, remains an open challenge at least in part because it is learned using unreliable reward models such as object detectors and vision-language models. 
We introduce \emph{Verifiable Visual Rewards} (VVR), the first framework for programmatically verifiable image rewards, and show that training on it generalizes to natural prompts. 
Each VVR task is a scene of geometric objects and relations among them, from which we derive both the prompt and a deterministic verifier, so tasks can be generated in any number and at any chosen complexity.
We release \bench, with 10,000 tasks over 32 constraint types, and \bench-Challenge, with 720 more complex tasks; 
the strongest model we evaluate---GPT-Image-2.5---solves 21.4\% of \bench-Challenge.
Using VVR scores as rewards for reinforcement learning (RLVVR) raises the accuracy of Stable Diffusion 3.5 Medium on \bench from 2.8\% to 28.3\% and demonstrates consistent easy-to-hard generalization. 
These gains extend to out-of-domain benchmarks, and mixing VVR into existing objectives further improves overall performance and human preference, motivating the adoption of VVR into standard image generation post-training recipes.

\end{abstract}

%% file: sec/1_intro.tex
\vspace{-2mm}
\section{Introduction}
\label{sec:intro}\vspace{-2mm}

Reinforcement learning with verifiable rewards has improved how precisely language models follow instructions: constraints such as \emph{use the word X at least three times} are checked by code and used directly as rewards~\citep{zhou2023ifeval,lambert2025tulu,pyatkin2025generalizing}, but image generation has no equivalent reward.
Text-to-image generators often fail to follow instructions precisely: given \emph{three red circles to the left of two blue squares}, they draw the wrong counts, colors, or positions, and fail more often as a prompt combines more requirements~\citep{ghosh2023geneval,huang2023t2icompbench,kamath2025geneval2}.
Post-training rewards for instruction following come from learned evaluators: preference models~\citep{kirstain2023pickscore,xu2023imagereward}, vision-language models (VLMs) that answer questions about the image~\citep{hu2023tifa,cho2023dsg,lin2024vqascore}, and object detectors~\citep{ghosh2023geneval}.
These evaluators make errors on the judgments that instruction following depends on~\citep{saxon2024t2iscorescore,wiles2025gecko,kajic2024geckonum,chen2025mjbench,kamath2025geneval2}, and policies trained on them exploit these errors~\citep{zhang2024overoptimization,kim2024confidence,hong2026rewardhacking}.

\begin{figure}[t]
\centering
\includegraphics[width=0.99\textwidth]{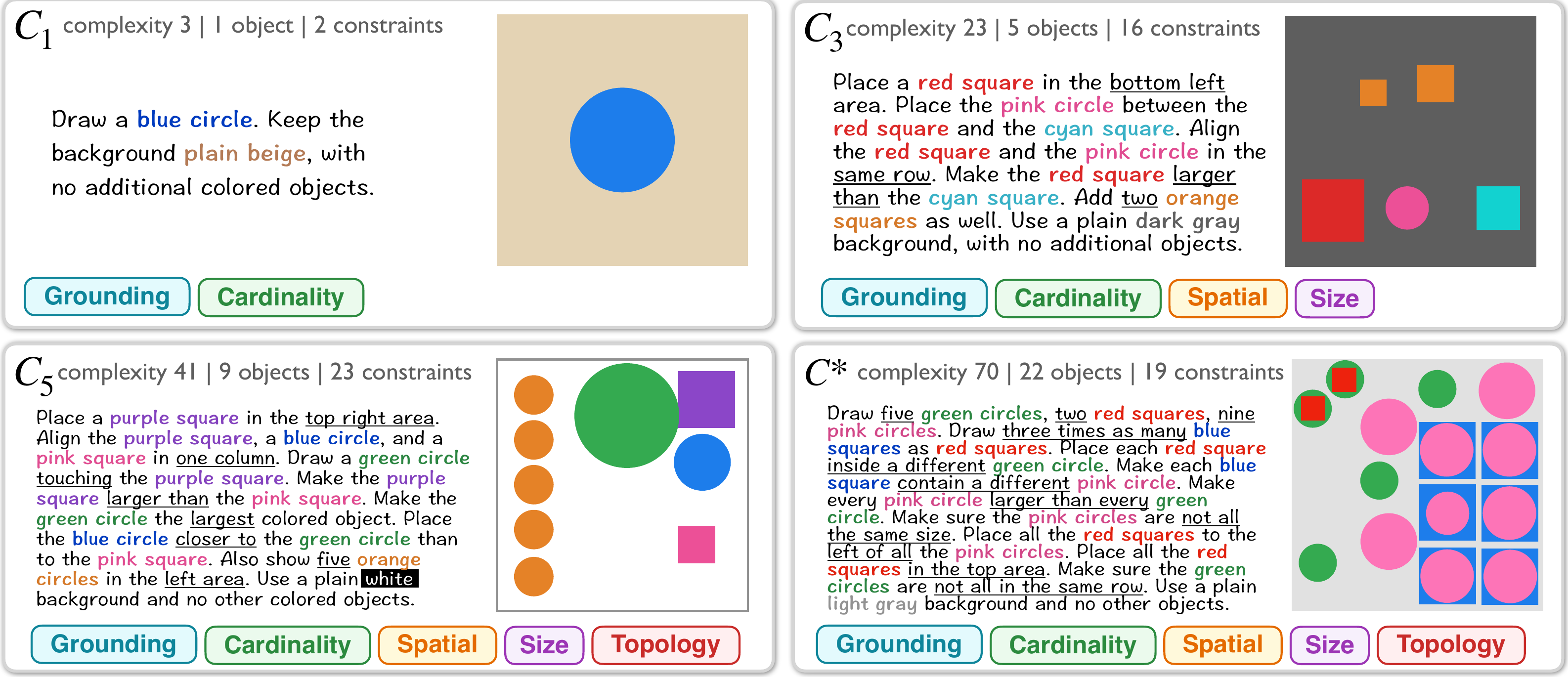}\vspace{-3mm}
\caption{Example \bench tasks from different complexity ranges ($C_1$--$C_5$) and Challenge ($C^{*}$).
Each panel shows the prompt, its complexity, the number of object instances, the number of constraints, the active constraint families, and a reference image that satisfies every constraint.}\vspace{-3mm}
\label{fig:vvr-task-examples}
\end{figure}

We introduce \emph{Verifiable Visual Rewards} (VVR), the first framework for programmatically verifiable image rewards, in which the generated image is scored deterministically by \textit{verifiers}: Python functions over pixels, with no learned detector, OCR system, embedding model, or VLM.
VVR covers instructions with clear, objective requirements combining color, count, shape, and spatial relations (Figure~\ref{fig:vvr-task-examples}).
Unlike constraints in text instruction following, which govern mostly separate properties of the output (e.g., length, keyword, format) and can be excluded pair by pair \citep{pyatkin2025generalizing},
visual constraints lead to more complicated compatibility conflicts. 
For example, in \textit{``\textcolor{wrong}{A contains B, B contains C, and C contains A},''} every pair of constraints can be satisfiable, but the three together are not.
Therefore, we propose a generator that guarantees constraint satisfiability under compositions, and compose natural-language instructions from the valid constraint sets. 
With our generator and constraint taxonomy, new VVR tasks can be generated in any number and at any chosen complexity, for evaluation or for training. 
Program verifiers of each VVR constraint also enable fine-grained diagnosis of generator capability over different types of instructions.

\textbf{\bench.}\hspace{2mm}
We instantiate VVR with colored geometric shapes and 46 constraint types over counts, attributes, and spatial relations and release \textbf{\bench}, with 10,000 tasks across five complexity ranges, and \textbf{\bench-Challenge}, with 720 more complex tasks to discriminate among frontier models (\S\ref{sec:vvr-bench}).
The strongest open-weight model, FLUX.2-dev, achieves 19.2\% accuracy on \bench, and the strongest model overall, GPT-Image-2.5-Sunburst, solves 21.4\% of \bench-Challenge.
Failures concentrate in the \CardinalityTag{Cardinality} (e.g., \textit{``twice as many A as B''}) and \TopologyTag{Topology} (e.g., \textit{``each A is inside a different B''}) constraint families. 
The benchmark can be updated with higher complexity as frontier models evolve. 

\textbf{RLVVR.}\hspace{2mm}
We use the verifiable VVR scores as rewards for reinforcement learning (RLVVR) to train image generators to follow instructions precisely.
To study how training complexity affects generalization, we procedurally generate two training corpora: VVR-Easy contains only low complexity tasks of at most one constraint family, and VVR-Matched matches the \bench distribution. 
We show that 
1) training on easy distribution generalizes to harder tasks, 
2) training on harder tasks teaches compositionality, 
3) training on colored shapes transfer to out-of-domain natural prompts to improve position and counting, 
and, most importantly, 
4) mixing VVR with existing post-training objectives (e.g., GenEval2, OCR, PickScore) improves general benchmark performance and human preference, 
motivating the adoption of VVR into standard image generation post-training recipes. 
Our contributions are:\vspace{-2mm}

\begin{enumerate}[leftmargin=*,itemsep=0pt,topsep=0pt]
\item VVR, the first framework for programmatically verifiable image rewards, whose generator composes constraints on color, count, shape, and spatial relations into satisfiable instructions at any chosen complexity, each constraint checked by its own verifier (\S\ref{sec:method}).
\item \bench and \bench-Challenge, benchmark that reveals capability gap of image generators to follow instructions,
on which even frontier image generators fail most complex tasks, with failures tracable to specific constraint types (\S\ref{sec:vvr-bench}).
\item RLVVR, reinforcement learning with VVR rewards, which improves precise instruction following on tasks harder than those seen in training, transfers from synthetic scenes to natural prompts, and, mixed with existing post-training objectives, improves general benchmark performance and human preference (\S\ref{sec:posttraining}).
\end{enumerate}

%% file: sec/2_methods.tex
\section{Verifiable Visual Rewards}
\label{sec:method}\vspace{-1mm}

\subsection{VVR Task Representation}
\label{sec:representation}\vspace{-1mm}

A VVR task specifies the requirements of a scene of colored shapes:\vspace{-4mm}

\begin{equation}\label{eq:representation}
    s=(\mathcal{G},\mathcal{B},\mathcal{A},\mathcal{F},p).
\end{equation}
Here $\mathcal{G}$ is a set of object groups, $\mathcal{B}$ is a set of background constraints, $\mathcal{A}$ is a set of active constraints on the object groups, $\mathcal{F}$ is a set of forbidden-content constraints, and $p$ is the natural-language instruction.
In \bench, $\mathcal{B}$ and $\mathcal{F}$ are the same in every task: a plain background color and $\mathcal{F}=\{$\texttt{no\_unrequested\_objects}$\}$, so we focus the rest of the section on the constraints in $\mathcal{A}$.

Each constraint in $\mathcal{A}$ is instantiated with a constraint \emph{type} from the constraint library and one or more object groups.
Each constraint type has a predefined number of object groups that it operates on, and a set of supported values (Table~\ref{tab:vvr-taxonomy}).
For example, \texttt{exact\_count(g1;\,3)} is a unary constraint that requires group \texttt{g1} to contain three objects, and \texttt{left\_of(g1,g2)} requires group \texttt{g1} to appear left of group \texttt{g2}.
Every group has exactly one color constraint and one shape constraint; all other constraints are optional.
Appendix~\ref{sec:representation-example} shows a complete task.

A valid task must satisfy three desiderata: 
\textbf{1) Well-formed:}
every constraint uses a defined type with supported parameter values, such as one of eight colors or three shapes, and refers to as many object groups in $\mathcal{G}$ as its type requires;
\textbf{2) Jointly satisfiable:}
at least one placement and sizing of the specified objects satisfies all constraints in $\mathcal{B}$, $\mathcal{A}$, and $\mathcal{F}$ simultaneously;
and
\textbf{3) Faithfully expressed:}
$p$ states every constraint in $\mathcal{B}$, $\mathcal{A}$, and $\mathcal{F}$ without any addition or omission.

\input{tables/vvr_verifier_taxonomy}

\subsection{Task generation}
\label{sec:task-generation}

We now walk through the stages of the VVR generator that produces well-formed, jointly satisfiable, and faithfully expressed tasks. Appendix~\ref{app:vvr_generator} provides an example generation and validation details. 
\begin{enumerate}[itemsep=0pt,topsep=0pt,leftmargin=15pt]
    \item It first creates a \emph{scene} by sampling background constraints $\mathcal{B}$ and objects. It randomly assigns every object a color, shape, position, and size, forming object groups $\mathcal{G}$, and samples forbidden-content constraints $\mathcal{F}$ that no object in the scene violates.
    \item For each constraint type in the library, it lists all object group tuples with size corresponding to the type's arity. The constraint type and its input tuple form an \emph{instantiated constraint}. 
    \item All instantiated constraints that are true under the constructed scene, checked by program verifiers, form \textit{satisfiable constraint set} $\mathcal{A}^*$.
    From $\mathcal{A}^*$, multiple valid active constraint sets $\mathcal{A}$ can be sampled such that $\mathcal{A}\subseteq\mathcal{A}^*$.
    \item A template $\tau$ with phrasing variants transforms each task requirement into natural language, $p=\tau(\mathcal{G},\mathcal{B},\mathcal{A},\mathcal{F})$, forming $s=(\mathcal{G},\mathcal{B},\mathcal{A},\mathcal{F},p)$. 
\end{enumerate}

Every constraint instantiates a library type on a tuple of sampled groups whose length equals the type's arity, so every task is \textbf{well-formed} by construction.
The scene satisfies $\mathcal{B}$, $\mathcal{F}$, and every constraint in $\mathcal{A}^*$, so it satisfies the task formed with any $\mathcal{A}\subseteq\mathcal{A}^*$, making the task \textbf{jointly satisfiable}.
Finally, in the template $\tau$, every requirement of $s$ has a fixed phrase in $p$, and every phrase in $p$ comes from a requirement of $s$, guaranteeing \textbf{expression faithfulness}. 

\vspace{-2mm}
\paragraph{Structural complexity estimates task difficulty.} 
We define the structural complexity of a task as $C(s)=\sum_{a\in\mathcal{A}}c(a;s)$, 
where $c(a;s)$ is the complexity contribution of one constraint $a$ in task $s$.
$c(a;s)$ follows a fixed rule for each constraint type and grows with the number of object instances that $a$ evaluates in $s$.
Appendix~\ref{app:constraint-library} gives the complexity contribution of every constraint type.
In the specific instantiation of tasks that produces datasets in Table~\ref{tab:vvr-sampler-instances}, each task has one background-color constraint and one forbidden-content constraint,
so the constraints in $\mathcal{B}$ and $\mathcal{F}$ are excluded.

\vspace{-2mm}
\paragraph{Datasets.}
Given a target distribution $\mathcal{T}$ over constraint families and structural complexity range, the generator can retain task candidates to fit $\mathcal{T}$. Thus datasets can be built to evaluate or learn specific constraint types at specified difficulty. VVR datasets used by this paper and their complexity distribution are listed in Table~\ref{tab:vvr-sampler-instances}.

\subsection{Deterministic, reference-free constraint verification}
\label{sec:verification}

VVR is an open-ended image generation task, where any image that satisfies all constraints receives full credit, so the verifier has to be reference-free. It takes in the generated RGB image $x$ and the formal constraints $(\mathcal{G},\mathcal{B},\mathcal{A},\mathcal{F})$ of the task and produces a correctness decision.

\vspace{-2mm}
\paragraph{Object extraction from pixels.}
VVR first extracts candidate objects from the generated image using deterministic pixel-level operations.
1) It produces a binary mask for each supported color, with fixed hue and contrast thresholds.
2) Connected-component analysis assigns the same label to foreground pixels connected by a path of edge- or corner-adjacent pixels; each labeled region is a candidate object.
3) Fixed contour measurements classify each candidate into one of the supported shapes
(circle, square, or triangle) based on its aspect ratio, bounding-box coverage, and convexity.
4) Candidates are then matched to object groups by the color and shape constraints of each group.
Appendix~\ref{sec:pixel-object-details} visualizes the extraction pipeline, including the color map and shape classifier,
as well as the handling of ambiguous colors, irregular contours, fragmented objects, and blurred boundaries. 

\vspace{-2mm}
\paragraph{Constraint verifier library.}
The verifier library $\mathcal{V}$ contains one \emph{program verifier} for each constraint type: a Python function that applies the type's requirement to the extracted objects.
Verifier decisions use fixed comparisons of object counts, positions, extents, or boundary distances.
For a constraint $a\in\mathcal{B}\cup\mathcal{A}\cup\mathcal{F}$, the verifier $v_a$ %
returns a pass-or-fail decision $d_a(x,s)\in\{0,1\}$ and a partial-credit score $q_a(x,s)\in[0,1]$.

\begin{wrapfigure}{r}{0.47\textwidth}\vspace{-8.3mm}
\begin{lstlisting}[
    language=Python,
    basicstyle=\ttfamily\scriptsize,
    frame=single,
    framesep=4pt,
    columns=fullflexible,
    showstringspaces=false
]
def left_of(g1, g2, m):
    g1x = np.mean([c.centroid[0] for c in g1])
    g2x = np.mean([c.centroid[0] for c in g2])
    delta = g2x - g1x
    partial = np.clip(delta / max(m, 1), 0, 1)
    return delta >= m, partial
\end{lstlisting}\vspace{-3mm}
\end{wrapfigure}

Consider the constraint 
$a=\texttt{left\_of(g1,g2)}$, 
its program verifier
computes $\delta$, the mean horizontal centroid coordinate of \texttt{g2} minus that of \texttt{g1}, and compares it with a separation margin $m$.
It returns two values: 
1) the decision $d_a$, which passes when \texttt{g1} lies to the left of \texttt{g2} by at least the margin; 
and 
2) the partial-credit score $q_a=\min(1,\max(0,\delta/m))$, the fraction of the required separation that the image achieves.
The score is 0 when \texttt{g1} is at or to the right of \texttt{g2}, rises linearly as \texttt{g1} moves left, and reaches 1 at the margin, where the decision also passes.
Appendix~\ref{app:program-verifiers} gives the verifier code for every constraint in an example task, Appendix~\ref{sec:verifier-validation-details} provides verifier validation details.

\vspace{-2mm}
\paragraph{Scores.}
A generated image succeeds only if every constraint in $\mathcal{B}$, $\mathcal{A}$, and $\mathcal{F}$ passes:\vspace{-5mm}

\begin{equation}\label{eq:accuracy}
    r_{\mathrm{exact}}(x,s)= \prod_{a\in\mathcal{B}\cup\mathcal{A}\cup\mathcal{F}}d_a(x,s).
\end{equation}\vspace{-4mm}

\bench accuracy is the mean of $r_{\mathrm{exact}}$ across tasks.
For training, we design a dense reward $r_{\mathrm{dense}}\in[0,1]$ that gives partial credit through the verifier partial-credit scores:\vspace{-5mm}

\begin{equation}\label{eq:dense_reward}
r_{\mathrm{dense}}(x,s)=\psi(x,s)\sum_{a\in\mathcal{B}\cup\mathcal{A}\cup\mathcal{F}} w_a\,q_a(x,s).
\end{equation}\vspace{-4mm}

The weights $w_a$ are fixed by constraint type, and $\psi(x,s)\in[0,1]$ is a multiplicative penalty factor that prevents the model from exploiting any single easy-to-learn constraint while ignoring others~\citep{zhang2024overoptimization,hong2026rewardhacking}. Appendix~\ref{app:scores} provides more details on $w_a$ and $\psi$.

%% file: tables/vvr_verifier_taxonomy.tex
\begin{table}[t]
\centering
\small
\vspace{-3mm}
\caption{\small
Constraint library. VVR groups 46 constraint types into five families.
Appendix~\ref{app:constraint-library} lists all types and their supported values, and Appendix~\ref{app:program-verifiers} defines their verifiers.}
\label{tab:vvr-taxonomy}
\resizebox{\linewidth}{!}{
\begin{tabular}{lll
}
\toprule
Family & Visual property & Example constraint types \\
\midrule
\GroundingTag{Grounding} & Object identity and attribute binding
& Color; shape; color--shape binding \\
\CardinalityTag{Cardinality} & Quantities and count comparisons
& Exact count; equal, greater, or fewer counts; count ratios (X times as many) \\
\SpatialTag{Spatial} & Position and arrangement
& Image regions; relative order; alignment; grids; distance comparisons \\
\SizeTag{Size} & Relative visual extent
& Pairwise and groupwise size; within-group variation; extrema \\
\TopologyTag{Topology} & Contact and enclosure
& Touching; separation; containment; distinct containment \\
\bottomrule
\end{tabular}
}\vspace{-3mm}
\end{table}

%% file: sec/3_benchmark.tex
\section{\texorpdfstring{\bench}{VVRBench}}
\label{sec:vvr-bench}

\begin{wrapfigure}{r}{0.61\textwidth}\vspace{-7.5mm}
\centering
\small
\setlength{\tabcolsep}{5pt}
\captionof{table}{\small
Training and evaluation datasets produced by the VVR task generator.
Appendix~\ref{app:generation-datasets} provides details on target distribution.
}
\label{tab:vvr-sampler-instances}
\begin{tabular}{@{}llrl@{}}
\toprule
Dataset & Use & \hspace{-8mm}Size & Complexity \\
\midrule
\bench & evaluation & 10,000 & 3--48 \\
\bench-Fast & evaluation & 820 & 3--44, 20 each \\
\bench-Challenge & evaluation & 720 & 45--80, 20 each \\
VVR-Easy & training & 100,000 & $\leq$ 20 \\
VVR-Matched & training & 100,000 & $\sim$\bench \\
\bottomrule
\end{tabular}
\vspace{-2mm}
\end{wrapfigure}

\vspace{-2mm}
\paragraph{Benchmark splits.}
We evaluate on three benchmark splits (Table~\ref{tab:vvr-sampler-instances}).
\textbf{1)~\bench}, the main benchmark, contains 10,000 tasks of complexity 3 to 48, which we report in five ranges $C_1$ to $C_5$ of about 2,000 tasks each.
\textbf{2)~\bench-Fast} is an 820-task subset covering the same range with 20 tasks at each integer complexity for evaluating image APIs at a twelfth of the generation cost.
\textbf{3)~\bench-Challenge} contains 720 tasks of complexity 45 to 80 and adds 14 more difficult, group level constraint types, above the \bench range, to separate the strongest generators. 

\vspace{-2mm}
\paragraph{Models.}
We evaluate ten open-weight models: 
FLUX.2-dev~\citep{bfl2025flux2}, 
HunyuanImage-2.1~\citep{tencent2025hunyuanimage21}, 
Qwen-Image-2512~\citep{wu2025qwenimage,qwen2025qwenimage2512}, 
HiDream-I1-Full~\citep{cai2025hidream}, 
FLUX.1-dev and FLUX.1-schnell~\citep{bfl2024flux}, 
Stable Diffusion 3.5 Medium and Large~\citep{esser2024scaling,stabilityai2024sd35}, 
SDXL~\citep{podell2024sdxl}, and 
Sana 1.6B~\citep{xie2025sana}, 
and seven API based models: 
GPT-Image-2.5-Sunburst~\citep{openai2026gptimage25,openai2026gptimage25card}, 
GPT-Image-2~\citep{openai2026gptimage2}, 
GPT-Image-1-mini~\citep{openai2025gptimage1mini},
Gemini-3.1-Flash-Image~\citep{google2026nanobanana2}, 
Gemini-3.1-Flash-Lite-Image~\citep{google2026flashliteimage}, 
Gemini-3-Pro-Image~\citep{google2025gemini3proimage}, and
Gemini-2.5-Flash-Image~\citep{google2025gemini25flashimage}.
Appendix~\ref{sec:evaluation-details} gives the additional evaluation details.

\vspace{-2mm}
\subsection{Precise instruction following is far from solved}
\label{sec:bench-accuracy}

\input{tables/vvr_bench_foundation_models}

\begin{figure*}[t]
\begin{minipage}[t]{0.42\textwidth}
\vspace{-0mm}
\centering
\includegraphics[width=\linewidth]{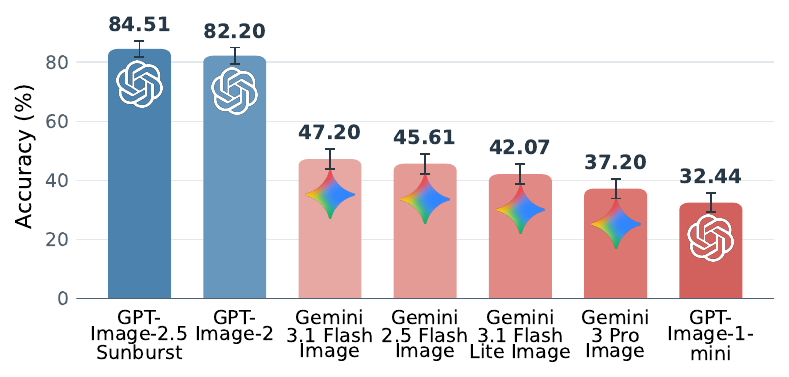}\vspace{-8mm}
\captionof{figure}{\small API models on \bench-Fast.}
\label{fig:vvr-bench-core-api-820}
\end{minipage}\hfill
\begin{minipage}[t]{0.55\textwidth}
\vspace{0pt}
\input{tables/vvr_bench_challenge_models}
\end{minipage}
\end{figure*}

As shown in Table~\ref{tab:vvr-bench-foundation-models}, the strongest open-weight model, FLUX.2-dev, solves 19.15\% of \bench tasks, and only 2.24\% in the high complexity bin $C_5$. Every other open-weight model solves less than 19\%.
GPT-Image-2 solves 86.86\% of all tasks, but its accuracy falls from 97.65\% in $C_1$ to 65.72\% in $C_5$.
On \bench-Fast (Figure~\ref{fig:vvr-bench-core-api-820}), GPT-Image-2.5-Sunburst and GPT-Image-2 solve 84.51\% and 82.20\% of the tasks, respectively, leading other API models by a large margin (exact scores in Appendix~\ref{sec:vvr-core820-details}).

\vspace{-2mm}
\paragraph{\bench-Challenge separates frontier models.}

With tasks in the complexity range of 3--44, \bench-Fast barely separates the strongest frontier models,
GPT-Image-2.5-Sunburst and GPT-Image-2, 
with a 2.3-points margin. 
Therefore, we create \bench-Challenge by sampling tasks from the uniform complexity distribution of 45--80 over a wider range of constraints
using the VVR generator (Appendix~\ref{app:generation-datasets}).
As shown in Table~\ref{tab:vvr-bench-challenge-models}, 
\bench-Challenge  discriminates among frontier models and exposes new failure modes.
GPT-Image-2.5-Sunburst solves 21.39\% of Challenge tasks, twice the 10.28\% of GPT-Image-2, and its accuracy falls from 31.67\% at complexity 45--56 to 24.58\% at 57--68 and 7.92\% at 69--80.
Every other model solves at most 8\% of \bench-Challenge, suggesting that there is still large room for improvement.
Interestingly, we observe occasional abstention behaviors from all Gemini models, stating the instruction is unsatisfiable, demonstrating failure in spatial reasoning (Appendix~\ref{sec:no-image-responses}).

The uniform drop of model accuracy across increasing complexity bins validates the design of the structural complexity score as a model-independent heuristic to generate tasks with controlled difficulty.
Appendix~\ref{sec:complexity-validation} provides more details on complexity as a predictor of failure. 

\begin{finding}
Open-weight generators fail most \bench tasks, and even the strongest API models lose accuracy sharply as complexity grows.
\end{finding}

\subsection{Failures concentrate in counting and object matching}
\label{sec:constraint-failures}

\begin{figure*}[t]
\centering
\includegraphics[width=\textwidth]{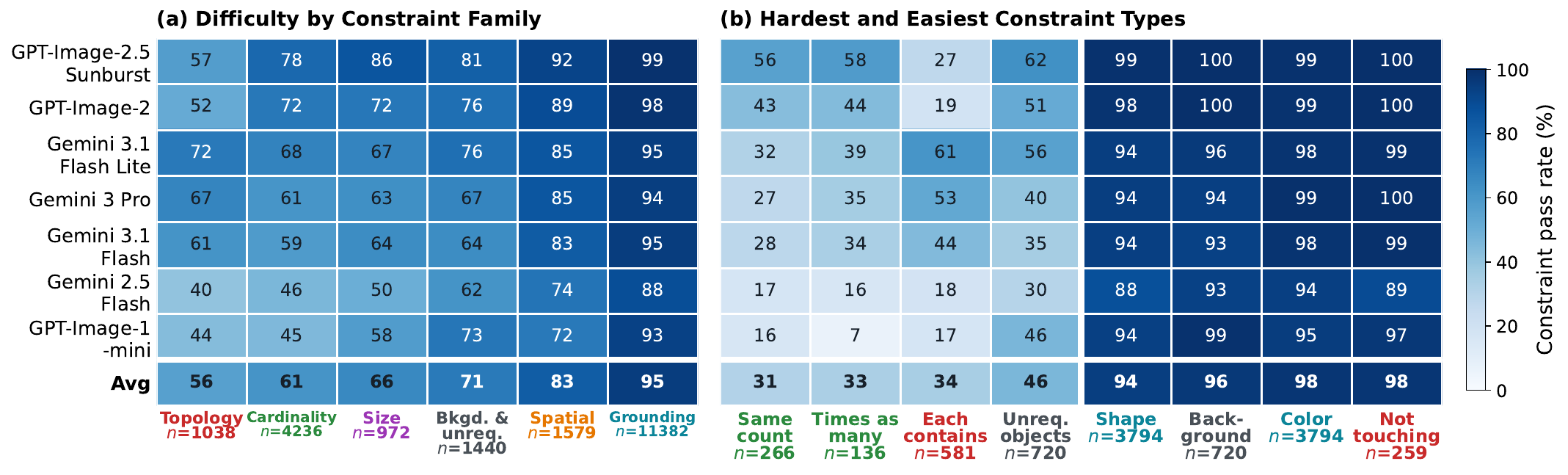}
\vspace{-8mm}
\caption{\small
Pass rates of individual constraint for API models on \bench-Challenge, by family (left) and for the four constraint types with the lowest and the four with the highest average pass rates among those with at least 100 checks per model (right).
}\vspace{-3mm}
\label{fig:vvr-api-atomic-capability}
\end{figure*}

The binary pass-fail score (Eq.~\ref{eq:accuracy}) is composed of individual verifier decisions from each of the active constraints in each task. Figure~\ref{fig:vvr-api-atomic-capability} presents the constraint-level pass rate of the API models on \bench-Challenge. 
Organized by constraint families (Table~\ref{tab:vvr-taxonomy}), 95\% of \GroundingTag{Grounding} constraints are satisfied, %
while only 56\% of \TopologyTag{Topology} constraints are rendered, averaged across models.
The hardest constraint types concern counts or relations across object groups: \CardinalityTag{same count} passes in 31\% of checks, \CardinalityTag{times as many} in 33\%, and \TopologyTag{each contains}, which requires each object of one group to contain a different object of another group, in 34\%.

Appendix~\ref{sec:api-atomic-details} gives the pass rate of every constraint type, 
Appendix~\ref{sec:family-presence-details} correlates each family to accuracy across all models at matched complexity, and Appendix~\ref{sec:failure-examples} shows typical failures in which a model adds objects that the prompt excludes.

\begin{finding}
Generators satisfy requirements on individual objects and pairs but fail requirements that constrain whole sets of objects.
\end{finding}

%% file: tables/vvr_bench_foundation_models.tex
\begin{table*}[t]
\centering
\scriptsize
\setlength{\tabcolsep}{2.5pt}
\caption{\small Accuracy (\%) on the 10,000 \bench tasks, overall and by complexity range.
$C_1$ to $C_5$ split the tasks by structural complexity $C(s)$ into five ranges of about 2,000 tasks each: 3--16, 16--21, 21--26, 26--31, and 31--48.
Accuracy generally falls with complexity.
GPT-Image-2 drops from 97.65\% in $C_1$ to 65.72\% in $C_5$, and no open weight model exceeds 20\% overall.
}
\label{tab:vvr-bench-foundation-models}
\begin{tabular}{lrrrrrr}
\toprule
Model & Accuracy (\%) $\uparrow$ & $C_1$ & $C_2$ & $C_3$ & $C_4$ & $C_5$ \\
\midrule
GPT-Image-2 & \vvrbenchscore{86.86}{97.0}{\textbf{86.86}$_{\pm 0.68}$} & \vvrbenchscore{97.65}{99.0}{\textbf{97.65}$_{\pm 0.74}$} & \vvrbenchscore{98.18}{100.0}{\textbf{98.18}$_{\pm 0.70}$} & \vvrbenchscore{91.00}{98.0}{\textbf{91.00}$_{\pm 1.34}$} & \vvrbenchscore{81.83}{96.0}{\textbf{81.83}$_{\pm 1.75}$} & \vvrbenchscore{65.72}{93.9}{\textbf{65.72}$_{\pm 2.10}$} \\
GPT-Image-1-mini & \vvrbenchscore{26.40}{88.9}{26.40$_{\pm 0.87}$} & \vvrbenchscore{74.06}{94.9}{74.06$_{\pm 1.93}$} & \vvrbenchscore{36.78}{90.9}{36.78$_{\pm 2.18}$} & \vvrbenchscore{12.47}{76.8}{12.47$_{\pm 1.53}$} & \vvrbenchscore{4.70}{60.6}{4.70$_{\pm 1.02}$} & \vvrbenchscore{2.39}{42.4}{2.39$_{\pm 0.76}$} \\
FLUX.2-dev & \vvrbenchscore{19.15}{82.8}{19.15$_{\pm 0.78}$} & \vvrbenchscore{49.86}{92.9}{49.86$_{\pm 2.15}$} & \vvrbenchscore{24.26}{86.9}{24.26$_{\pm 1.96}$} & \vvrbenchscore{12.42}{75.8}{12.42$_{\pm 1.52}$} & \vvrbenchscore{5.91}{63.6}{5.91$_{\pm 1.12}$} & \vvrbenchscore{2.24}{41.4}{2.24$_{\pm 0.74}$} \\
HunyuanImage-2.1 & \vvrbenchscore{18.79}{81.8}{18.79$_{\pm 0.78}$} & \vvrbenchscore{45.53}{91.9}{45.53$_{\pm 2.15}$} & \vvrbenchscore{19.64}{84.8}{19.64$_{\pm 1.83}$} & \vvrbenchscore{14.84}{77.8}{14.84$_{\pm 1.63}$} & \vvrbenchscore{8.61}{68.7}{8.61$_{\pm 1.31}$} & \vvrbenchscore{4.29}{57.6}{4.29$_{\pm 0.98}$} \\
Qwen-Image-2512 & \vvrbenchscore{5.79}{61.6}{5.79$_{\pm 0.47}$} & \vvrbenchscore{19.16}{83.8}{19.16$_{\pm 1.75}$} & \vvrbenchscore{5.87}{62.6}{5.87$_{\pm 1.14}$} & \vvrbenchscore{2.11}{39.4}{2.11$_{\pm 0.73}$} & \vvrbenchscore{1.10}{34.3}{1.10$_{\pm 0.56}$} & \vvrbenchscore{0.15}{22.7}{0.15$_{\pm 0.29}$} \\
HiDream-I1-Full & \vvrbenchscore{4.22}{56.6}{4.22$_{\pm 0.41}$} & \vvrbenchscore{16.47}{79.8}{16.47$_{\pm 1.65}$} & \vvrbenchscore{3.06}{49.5}{3.06$_{\pm 0.87}$} & \vvrbenchscore{0.80}{31.3}{0.80$_{\pm 0.50}$} & \vvrbenchscore{0.20}{24.7}{0.20$_{\pm 0.31}$} & \vvrbenchscore{0.00}{8.1}{0.00$_{\pm 0.19}$} \\
FLUX.1-dev & \vvrbenchscore{3.87}{52.5}{3.87$_{\pm 0.40}$} & \vvrbenchscore{15.51}{78.8}{15.51$_{\pm 1.62}$} & \vvrbenchscore{2.18}{40.4}{2.18$_{\pm 0.75}$} & \vvrbenchscore{0.96}{33.3}{0.96$_{\pm 0.53}$} & \vvrbenchscore{0.15}{22.7}{0.15$_{\pm 0.29}$} & \vvrbenchscore{0.00}{8.1}{0.00$_{\pm 0.19}$} \\
FLUX.1-schnell & \vvrbenchscore{2.88}{48.5}{2.88$_{\pm 0.35}$} & \vvrbenchscore{11.96}{73.7}{11.96$_{\pm 1.46}$} & \vvrbenchscore{1.66}{38.4}{1.66$_{\pm 0.67}$} & \vvrbenchscore{0.25}{26.3}{0.25$_{\pm 0.34}$} & \vvrbenchscore{0.10}{20.7}{0.10$_{\pm 0.26}$} & \vvrbenchscore{0.00}{8.1}{0.00$_{\pm 0.19}$} \\
SD3.5 Medium & \vvrbenchscore{2.81}{47.5}{2.81$_{\pm 0.34}$} & \vvrbenchscore{12.01}{74.7}{12.01$_{\pm 1.47}$} & \vvrbenchscore{1.19}{37.4}{1.19$_{\pm 0.59}$} & \vvrbenchscore{0.35}{28.3}{0.35$_{\pm 0.37}$} & \vvrbenchscore{0.05}{18.7}{0.05$_{\pm 0.23}$} & \vvrbenchscore{0.00}{8.1}{0.00$_{\pm 0.19}$} \\
SD3.5 Large & \vvrbenchscore{2.45}{43.4}{2.45$_{\pm 0.32}$} & \vvrbenchscore{10.47}{71.7}{10.47$_{\pm 1.39}$} & \vvrbenchscore{1.14}{36.4}{1.14$_{\pm 0.58}$} & \vvrbenchscore{0.20}{24.7}{0.20$_{\pm 0.32}$} & \vvrbenchscore{0.05}{18.7}{0.05$_{\pm 0.23}$} & \vvrbenchscore{0.00}{8.1}{0.00$_{\pm 0.19}$} \\
SDXL 1.0 & \vvrbenchscore{0.02}{17.2}{0.02$_{\pm 0.05}$} & \vvrbenchscore{0.10}{20.7}{0.10$_{\pm 0.25}$} & \vvrbenchscore{0.00}{8.1}{0.00$_{\pm 0.20}$} & \vvrbenchscore{0.00}{8.1}{0.00$_{\pm 0.19}$} & \vvrbenchscore{0.00}{8.1}{0.00$_{\pm 0.19}$} & \vvrbenchscore{0.00}{8.1}{0.00$_{\pm 0.19}$} \\
Sana 1.6B & \vvrbenchscore{0.00}{8.1}{0.00$_{\pm 0.04}$} & \vvrbenchscore{0.00}{8.1}{0.00$_{\pm 0.18}$} & \vvrbenchscore{0.00}{8.1}{0.00$_{\pm 0.20}$} & \vvrbenchscore{0.00}{8.1}{0.00$_{\pm 0.19}$} & \vvrbenchscore{0.00}{8.1}{0.00$_{\pm 0.19}$} & \vvrbenchscore{0.00}{8.1}{0.00$_{\pm 0.19}$} \\
\bottomrule
\end{tabular}
\end{table*}

%% file: tables/vvr_bench_challenge_models.tex
\centering
\setlength{\tabcolsep}{2pt}
\vspace{-3mm}
\captionof{table}{\small
\bench-Challenge acc. (\%).
The best model solves 21.39\% overall and 7.92\% at complexity 69 to 80.}
\label{tab:vvr-bench-challenge-models}
\resizebox{\linewidth}{!}{%
\begin{tabular}{lrrrr}
\toprule
Model & Accuracy $\uparrow$ & 45 to 56 & 57 to 68 & 69 to 80 \\
\midrule
GPT-Image-2.5-Sunburst & \vvrbenchscore{21.39}{85.9}{\textbf{21.39}$_{\pm 3.14}$} & \vvrbenchscore{31.67}{89.9}{\textbf{31.67}$_{\pm 6.13}$} & \vvrbenchscore{24.58}{87.9}{\textbf{24.58}$_{\pm 5.82}$} & \vvrbenchscore{7.92}{67.2}{\textbf{7.92}$_{\pm 4.12}$} \\
GPT-Image-2 & \vvrbenchscore{10.28}{70.7}{10.28$_{\pm 2.43}$} & \vvrbenchscore{17.50}{80.8}{17.50$_{\pm 5.31}$} & \vvrbenchscore{10.83}{72.7}{10.83$_{\pm 4.57}$} & \vvrbenchscore{2.50}{45.5}{2.50$_{\pm 2.85}$} \\
Gemini-3.1-Flash-Lite-Image & \vvrbenchscore{7.36}{65.7}{7.36$_{\pm 2.14}$} & \vvrbenchscore{10.00}{69.7}{10.00$_{\pm 4.45}$} & \vvrbenchscore{4.17}{55.1}{4.17$_{\pm 3.33}$} & \vvrbenchscore{7.92}{67.2}{\textbf{7.92}$_{\pm 4.12}$} \\
Gemini-3-Pro-Image & \vvrbenchscore{4.58}{59.1}{4.58$_{\pm 1.78}$} & \vvrbenchscore{7.08}{64.6}{7.08$_{\pm 3.97}$} & \vvrbenchscore{4.17}{55.1}{4.17$_{\pm 3.33}$} & \vvrbenchscore{2.50}{45.5}{2.50$_{\pm 2.85}$} \\
Gemini-3.1-Flash-Image & \vvrbenchscore{3.89}{53.5}{3.89$_{\pm 1.67}$} & \vvrbenchscore{3.33}{50.5}{3.33$_{\pm 3.11}$} & \vvrbenchscore{3.75}{51.5}{3.75$_{\pm 3.22}$} & \vvrbenchscore{4.58}{59.1}{4.58$_{\pm 3.44}$} \\
Gemini-2.5-Flash-Image & \vvrbenchscore{1.11}{35.4}{1.11$_{\pm 1.07}$} & \vvrbenchscore{2.50}{45.5}{2.50$_{\pm 2.85}$} & \vvrbenchscore{0.42}{29.8}{0.42$_{\pm 1.91}$} & \vvrbenchscore{0.42}{29.8}{0.42$_{\pm 1.91}$} \\
GPT-Image-1-mini & \vvrbenchscore{0.28}{27.3}{0.28$_{\pm 0.73}$} & \vvrbenchscore{0.83}{32.3}{0.83$_{\pm 2.15}$} & \vvrbenchscore{0.00}{8.1}{0.00$_{\pm 1.58}$} & \vvrbenchscore{0.00}{8.1}{0.00$_{\pm 1.58}$} \\
\bottomrule
\end{tabular}
}

%% file: sec/4_posttraining.tex
\section{RLVVR: VVR for Diffusion Post-Training}
\label{sec:posttraining}

VVR scores images with program verifiers, avoiding error propagation from unreliable learned evaluators, and 
is not limited to fixed prompt sets, so training can be scaled to any desired data size and difficulty distributions.
These properties allow us to improve image generation instruction following by using VVR as a reward in reinforcement learning (RLVVR). 
In this section, we post-train image generators with RLVVR to answer the following research questions:

\begin{itemize}[itemsep=0pt,topsep=0pt,leftmargin=28pt]
    \item[\textit{\textbf{RQ1.}}]Does RLVVR teach precise instruction following, and how does the complexity of the training tasks shape what is learned?
    \item[\textit{\textbf{RQ2.}}]Do the skills learned from synthetic scenes transfer to natural prompts beyond VVR?
    \item[\textit{\textbf{RQ3.}}]Is supervision from synthetic scenes complementary to existing post-training rewards?
\end{itemize}

\subsection{Experimental setup}
\label{sec:posttraining-setup}

\vspace{-1mm}
\paragraph{Data.} 
We generate two training corpora using the VVR generator (\S\ref{sec:task-generation}). \textbf{VVR-Easy} contains tasks that contain at most one constraint family and have complexity of at most 20, and \textbf{VVR-Matched} matches the \bench distribution (complexity 3--48). Each dataset contains 100K VVR tasks after decontamination from benchmark data (Table~\ref{tab:vvr-sampler-instances}).
For reward-mixture experiments, we train with GenEval2~\citep{kamath2025geneval2}, OCR~\citep{liu2025flow}, and a five-reward objective that combines GenEval \citep{ghosh2023geneval}, GenEval2, OCR, PickScore~\citep{kirstain2023pickscore}, and UnifiedReward~\citep{wang2025unifiedreward}.
Each of these objectives is trained alone and mixed with VVR-Easy, with equal number of prompts per objective.

\vspace{-3mm}
\paragraph{Training.}
We train Stable Diffusion 3.5 Medium~\citep{esser2024scaling,stabilityai2024sd35} with Flow-GRPO~\citep{liu2025flow}. 
Reward for each rollout is assigned by the scorer of the task objective that its prompt comes from. 
We use the VVR dense score $r_{\mathrm{dense}}$ for VVR prompts (Eq.~\ref{eq:dense_reward}).
We name each trained model after its training data. Appendix~\ref{sec:training-setup} reports training details.

\vspace{-3mm}
\paragraph{Evaluation.}
We evaluate trained models on \bench, GenEval, GenEval2, OCR, PickScore, HPSv2.1, CLIPScore, aesthetic score, ImageReward, HPSv3, and UnifiedReward (Appendix~\ref{sec:posttraining-eval-details}).

\subsection{RQ1: RLVVR teaches precise instruction following}
\label{sec:complexity-transfer}

\begin{figure*}[t]
\begin{minipage}[t]{0.52\textwidth}
\vspace{-0mm}
\centering
\includegraphics[width=\linewidth]{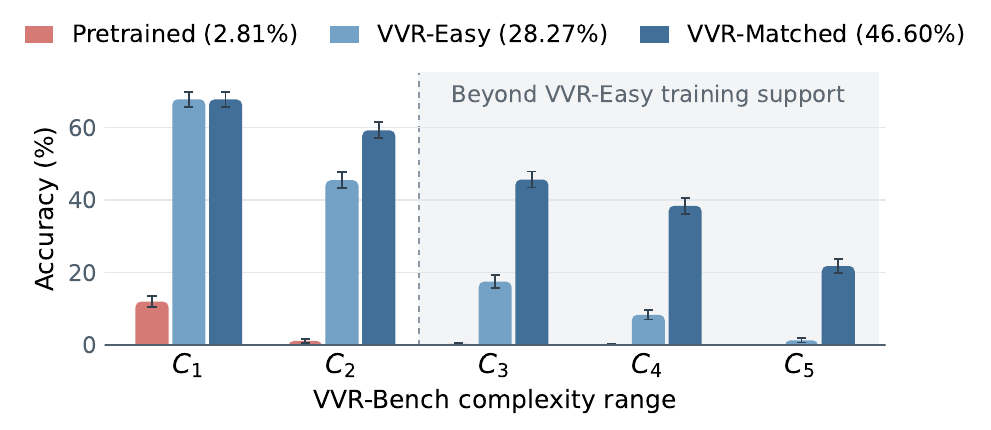}\vspace{-8mm}
\caption{\small
\bench accuracy by complexity range.
Shaded ranges lie above the complexity of every VVR-Easy training task, and VVR-Easy improves them.
Training on harder generated tasks (VVR-Matched) closes more of the gap.}\vspace{-4mm}
\label{fig:vvr-easy-complexity-transfer}
\end{minipage}\hfill
\begin{minipage}[t]{0.45\textwidth}
\vspace{0pt}
\includegraphics[width=\linewidth]{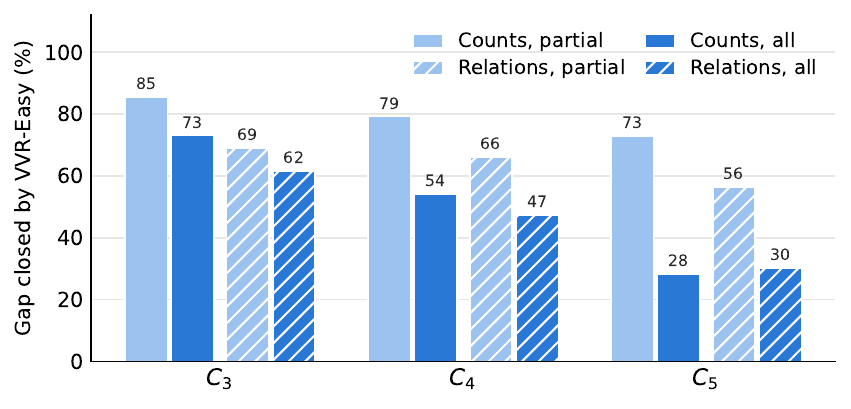}\vspace{-6.85mm}
\caption{\small VVR-Easy closes the gap between pretrained and VVR-Match more effectively on \cblock{156}{195}{239} partial scores (individual constraint) than \cblock{42}{120}{214} all-satisfy scores (compositionality) on both \emph{count} and \emph{relation} constraints.
}
\label{fig:vvr-partial-joint-gap}
\end{minipage}
\end{figure*}

Training on VVR-Easy raises \bench accuracy from 2.81\% to 28.27\% (Figure~\ref{fig:vvr-easy-complexity-transfer}).
Every task in $C_3$--$C_5$ is more complex than any VVR-Easy task, and on these ranges accuracy still rises by 17.16, 8.31, and 1.35 points.
Training on data from the benchmark distribution with VVR-Matched raises accuracy to 46.60\% overall and to 45.62\%, 38.39\%, and 21.82\% on $C_3$--$C_5$ (Appendix~\ref{sec:posttraining-vvr-bench-full}).

We separate how reliably a model satisfies individual constraints from how well it satisfies compositional requirements, using the two kinds of constraints that nearly every complex task contains: \textit{counts} and \textit{relations}. We compare the models' partial scores $q_a(x,s)$ on these constraints with how often they satisfy \textit{every} count or \textit{every} relation (Appendix~\ref{sec:partial-joint}).
On tasks outside of its training complexity range, VVR-Easy closes 79\% and 64\% of the gap between the pretrained model and VVR-Matched in the partial scores of counts and relations, respectively, but only 55\% and 46\% in how often all counts or all relations in a task are satisfied (Figure~\ref{fig:vvr-partial-joint-gap}).
Easy tasks thus make individual constraints reliable, and satisfying many constraints in the same image is learned from large scenes.

\begin{finding}
Training only on easy tasks makes individual constraints reliable, including on harder tasks, and training on large scenes teaches compositionality.
\end{finding}

\subsection{RQ2: Skills learned from synthetic scenes transfer to natural prompts}
\label{sec:natural-transfer}
\label{sec:human-preference}

Trained only on colored shapes, VVR-Easy improves over the pretrained reference on eight of ten non-VVR metrics, including GenEval by 0.113 and OCR by 0.111 (Table~\ref{tab:sd35-postfreeze-external}).
Human annotators confirm the transfer: 
VVR-Easy is preferred by annotators over the pretrained SD3.5-M on their generations from 160 natural prompts outside VVR with a win rate of 71.6\% with 83.8\% pairwise agreement (Table~\ref{tab:human-model-preference-final}).
Appendix~\ref{sec:human-study-details} reports annotation details. 

VVR-Easy also scores higher on 6 out of 9 natural prompt benchmarks than the model trained with the GenEval2 reward, whose training prompts name real objects---%
especially GenEval (0.729 vs. 0.688) and OCR (0.587 vs. 0.501).
Notably, the GenEval gain comes from position ($+0.150$), counting ($+0.103$), and color attribution ($+0.025$), all skills that VVR trains, while the single object, two object, and colors categories are comparable to the GenEval2-trained model.

\begin{finding}
Skills learned from synthetic VVR scenes transfer to natural prompts, especially in position and counting.
\end{finding}

\input{tables/sd35_postfreeze_external}

\input{tables/human_model_preference_final}

\subsection{RQ3: Supervision from synthetic scenes complements existing rewards}
\label{sec:mixture-results}

Mixed with GenEval2, VVR-Easy raises all ten non-VVR metrics in Table~\ref{tab:sd35-postfreeze-external}, including GenEval2 itself ($+0.025$).
The largest metric gains are in GenEval ($+0.030$), OCR ($+0.030$), HPSv3 ($+0.222$), and ImageReward ($+0.045$).
Human annotators prefer the mixture to GenEval2 alone on the 160 non-VVR natural prompts with a win rate of 58.6\% (Table~\ref{tab:human-model-preference-final}).
Combining with VVR-Matched, the dataset with more complex tasks and diverse constraint combinations, further raises performance and generalization on most natural prompts.

Mixed with OCR and with the five-reward objective, VVR-Easy raises eight of ten metrics each, with the largest gains in GenEval by 0.053 and ImageReward by 0.092 in the OCR mixture, and GenEval2 by 0.041 and HPSv3 by 0.307 in the five-reward mixture.
In these two mixtures, native OCR accuracy falls by 0.021 and 0.030, and in the five-reward mixture PickScore falls by 0.002, since each mixture trains on fewer prompts from the original sources.

\begin{finding}
Adding VVR tasks to existing post-training objectives improves human preference and most non-VVR metrics.
\end{finding}

\begin{figure*}[t]
\centering
\includegraphics[width=\textwidth]{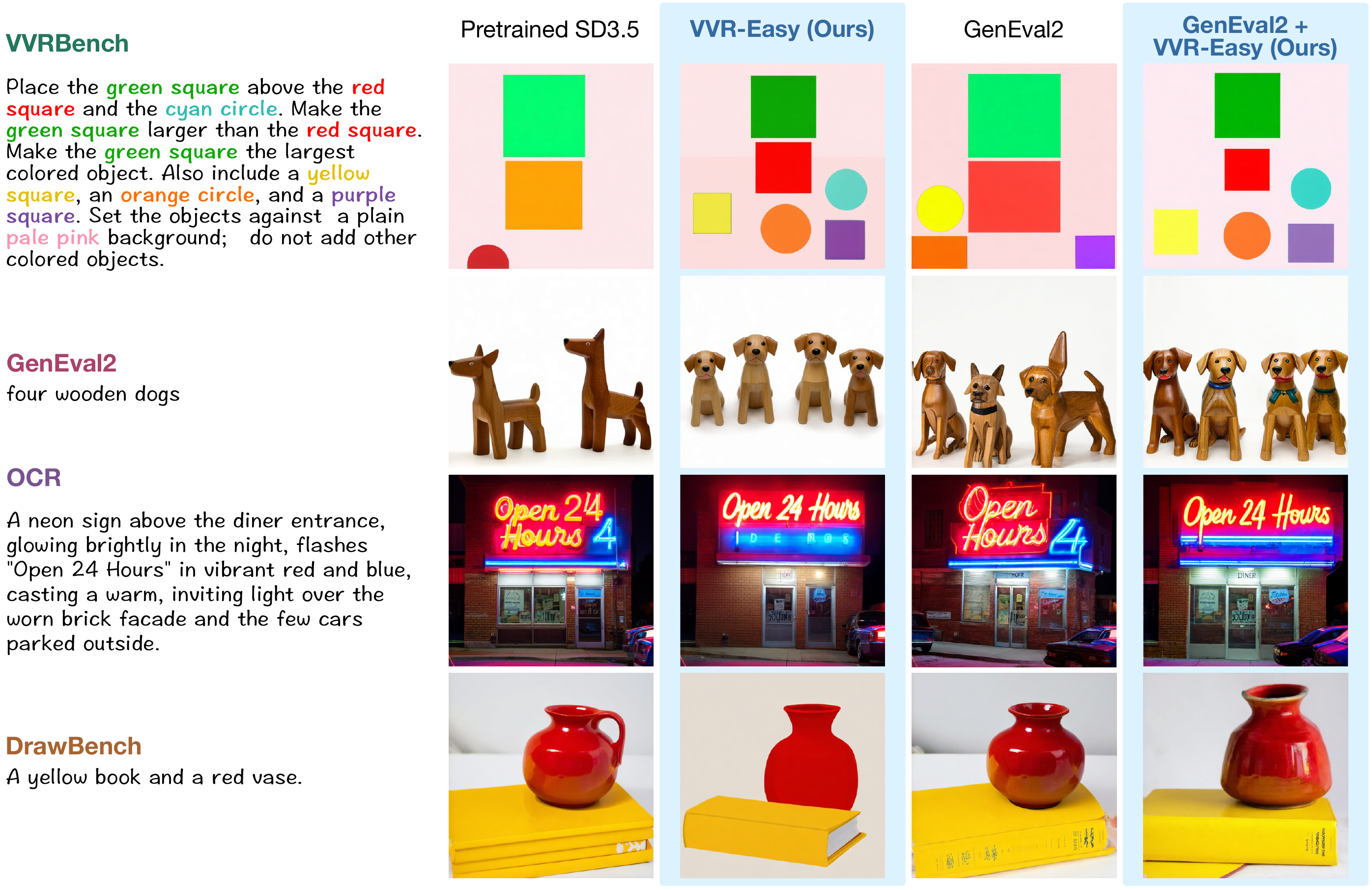}\vspace{-7mm}
\caption{\small
Example generations from the RLVVR and baseline models
from the same prompts and initial seed.
On the VVR prompt, the verifier accepts both RLVVR outputs and rejects both baselines.
On the DrawBench prompt, VVR-Easy renders the vase as a flat shape without shading, consistent with its lower aesthetic and HPSv2.1 scores; adding GenEval2 to VVR-Easy recovers shading and improves aesthetic and HPSv2.1.}\vspace{-1mm}
\label{fig:final-model-qualitative}
\end{figure*}

%% file: tables/sd35_postfreeze_external.tex
\begin{table}[t]
\vspace{-3mm}
\centering
\setlength{\tabcolsep}{2.7pt}
\caption{\small
RLVVR with VVR-Easy and reward mixtures transfers to most benchmarks and metrics.
}
\label{tab:sd35-postfreeze-external}
\newcommand{\vvrhi}[2]{\cellcolor{vvrHigh!#1!white}#2}
\newcommand{\vvrlo}[2]{\cellcolor{vvrLow!#1!white}#2}
\resizebox{\textwidth}{!}{%
\begin{tabular}{lrrrrrrrrrrr}
\toprule
Training reward & VVR & GenEval & GenEval2 & OCR & PickScore & HPSv2.1 & HPSv3 & CLIPScore & Aesthetic & ImageReward & UnifiedReward \\
\midrule
Pretrained & 0.028 & 0.616 & 0.237 & 0.476 & 0.841 & 0.300 & 7.689 & 0.956 & 5.517 & 0.929 & 0.636 \\
\hspace{1mm}$+$ VVR-Easy & 0.283 & 0.729 & 0.268 & 0.587 & 0.849 & 0.294 & 8.275 & 0.979 & 5.483 & 1.114 & 0.641 \\
\textbf{$\Delta$} & \vvrhi{26}{+0.255} & \vvrhi{30}{+0.113} & \vvrhi{23}{+0.031} & \vvrhi{30}{+0.111} & \vvrhi{30}{+0.008} & \vvrlo{30}{$-0.006$} & \vvrhi{30}{+0.586} & \vvrhi{30}{+0.023} & \vvrlo{30}{$-0.034$} & \vvrhi{30}{+0.185} & \vvrhi{30}{+0.005} \\
\midrule
GenEval2 & 0.039 & 0.688 & 0.454 & 0.501 & 0.848 & 0.297 & 8.289 & 0.974 & 5.523 & 1.110 & 0.635 \\
\hspace{1mm}$+$ VVR-Easy & 0.218 & 0.718 & 0.478 & 0.532 & 0.848 & 0.302 & 8.511 & 0.976 & 5.533 & 1.155 & 0.637 \\
\textbf{$\Delta$} & \vvrhi{18}{+0.180} & \vvrhi{8}{+0.030} & \vvrhi{18}{+0.025} & \vvrhi{8}{+0.030} & \vvrhi{2}{+0.0004} & \vvrhi{25}{+0.005} & \vvrhi{11}{+0.222} & \vvrhi{3}{+0.002} & \vvrhi{8}{+0.009} & \vvrhi{7}{+0.045} & \vvrhi{11}{+0.0019} \\
\hspace{1mm}$+$ VVR-Matched & 0.335 & 0.712 & 0.491 & 0.510 & 0.848 & 0.301 & 8.498 & 0.972 & 5.516 & 1.148 & 0.637 \\
\textbf{$\Delta$} & \vvrhi{30}{+0.296} & \vvrhi{6}{+0.024} & \vvrhi{28}{+0.038} & \vvrhi{2}{+0.009} & \vvrlo{1}{$-0.0002$} & \vvrhi{20}{+0.004} & \vvrhi{11}{+0.209} & \vvrlo{1}{$-0.001$} & \vvrlo{6}{$-0.007$} & \vvrhi{6}{+0.038} & \vvrhi{7}{+0.0012} \\
\midrule
OCR & 0.036 & 0.625 & 0.225 & 0.962 & 0.844 & 0.290 & 7.601 & 0.960 & 5.477 & 0.994 & 0.633 \\
\hspace{1mm}$+$ VVR-Easy & 0.247 & 0.678 & 0.252 & 0.941 & 0.846 & 0.290 & 7.765 & 0.969 & 5.488 & 1.087 & 0.636 \\
\textbf{$\Delta$} & \vvrhi{21}{+0.212} & \vvrhi{14}{+0.053} & \vvrhi{20}{+0.027} & \vvrlo{6}{$-0.021$} & \vvrhi{7}{+0.0018} & 0.000 & \vvrhi{8}{+0.164} & \vvrhi{12}{+0.009} & \vvrhi{11}{+0.012} & \vvrhi{15}{+0.092} & \vvrhi{14}{+0.0024} \\
\midrule
Five-reward & 0.048 & 0.739 & 0.342 & 0.856 & 0.850 & 0.294 & 8.137 & 0.974 & 5.510 & 1.125 & 0.640 \\
\hspace{1mm}$+$ VVR-Easy & 0.158 & 0.751 & 0.383 & 0.826 & 0.848 & 0.300 & 8.444 & 0.976 & 5.541 & 1.152 & 0.640 \\
\textbf{$\Delta$} & \vvrhi{11}{+0.110} & \vvrhi{3}{+0.012} & \vvrhi{30}{+0.041} & \vvrlo{8}{$-0.030$} & \vvrlo{9}{$-0.0024^\dagger$} & \vvrhi{30}{+0.006} & \vvrhi{16}{+0.307} & \vvrhi{3}{+0.002} & \vvrhi{27}{+0.031} & \vvrhi{4}{+0.027} & \vvrhi{2}{+0.0003} \\
\bottomrule
\end{tabular}%
}
\end{table}

%% file: tables/human_model_preference_final.tex
\begin{table*}[t]
\centering
\vspace{-3mm}
\setlength{\tabcolsep}{4pt}
\caption{Human preference win-rate (\%) for the RLVVR-trained model over its baseline.
}
\label{tab:human-model-preference-final}
\resizebox{\linewidth}{!}{
\begin{tabular}{lcccccc}
\toprule
VVR win rate vs. baseline & VVR & GenEval2 & GenEval & OCR & DrawBench & Outside VVR \\
\midrule
VVR-Easy vs.\ pretrained
& 93.3 & 78.8 & 62.9 & 69.6 & 75.0 & 71.6 \\
& {\scriptsize[86.7, 98.3]} & {\scriptsize[67.5, 88.8]} & {\scriptsize[50.4, 75.0]} & {\scriptsize[58.3, 80.4]} & {\scriptsize[65.4, 84.2]} & {\scriptsize[65.9, 77.1]} \\
\midrule
GenEval2 $+$ VVR-Easy vs.\ GenEval2
& 89.2 & 62.9 & 57.5 & 56.7 & 57.5 & 58.6 \\
& {\scriptsize[81.7, 95.4]} & {\scriptsize[50.8, 74.6]} & {\scriptsize[45.0, 69.6]} & {\scriptsize[44.6, 68.8]} & {\scriptsize[45.4, 69.2]} & {\scriptsize[52.6, 64.6]} \\
\bottomrule
\end{tabular}
}
\end{table*}

%% file: sec/5_related_work.tex
\section{Related Work}
\label{sec:related}

\textbf{Verifiable rewards.}
Verifiable rewards score language-model outputs with executable rules, such as exact-answer checks \citep{guo2025deepseekr1} and instruction-constraint checks \citep{zhou2023ifeval,lambert2025tulu}, and procedural environments generate such tasks at controlled difficulty \citep{stojanovski2025reasoninggym,liu2025synlogic,chen2025enigmata}.
\citet{johnson2017clevr} derive visual questions and their answers from generated scenes; VVR derives image-generation prompts and their constraints the same way.
For generated SVG and TikZ programs, rewards check the geometry of the rendered program~\citep{li2026geosvgrl} or compare its rendering with a reference image~\citep{rodriguez2025rlrf,belouadi2024detikzify}; VVR instead verifies generated pixels, with no program or reference image, and accepts any image that satisfies the constraints.

\textbf{Rewards for text-to-image post-training.}
Diffusion and flow models are post-trained with policy gradients \citep{black2024ddpo,fan2023dpok}, differentiable rewards \citep{xu2023imagereward,clark2024directly}, preference optimization~\citep{wallace2024diffusion}, and online reinforcement learning for flow models~\citep{liu2025flow,xue2025dancegrpo}.
Rewards that check the prompt rely on learned models: preference models \citep{kirstain2023pickscore,xu2023imagereward,wu2023human,wang2025unifiedreward}, or rules applied to the outputs of learned detectors and vision-language models.
\citet{liu2025flow} score GenEval detections~\citep{ghosh2023geneval} and OCR outputs, \citet{zhou2026spatialreward} combine detectors with a vision-language model, and \citet{huang2026alphagrpo} answer decomposed questions with a multimodal model.
Errors in these learned signals can be exploited during optimization~\citep{zhang2024overoptimization}.
The compressibility reward of \citet{black2024ddpo} needs no learned model but does not depend on the prompt.
RLVVR computes a prompt-specific reward from the generated pixels without a learned model, and it can be mixed with these objectives.

\textbf{Text-to-image evaluation.}
Text-to-image evaluation uses embedding and question-answering metrics \citep{hessel2021clipscore,hu2023tifa,cho2023dsg,lin2024vqascore} and prompt-alignment and compositional benchmarks \citep{saharia2022imagen,yu2022parti,ghosh2023geneval,huang2023t2icompbench,hu2024ella}, all of which score images with learned models.
\citet{kamath2025geneval2} replace the GenEval detector with a vision-language judge because detector scores diverged from human judgments on stronger generators.
\citet{wu2024conceptmix} and \citet{cho2024layoutbench} use synthetic visual concepts in their prompts but score the outputs with a detector or a VLM.
\bench scores every constraint exactly, with the same program verifiers that provide the RLVVR reward.

%% file: sec/6_conclusion.tex
\section{Conclusion}
\label{sec:conclusion}

In this paper, we introduce Verifiable Visual Rewards (VVR), where open-ended image generation can be scored by deterministic program verifiers to provide both evaluation feedback and post-training signals.
VVR tasks can be generated procedurally given any target distribution over constraint types and complexity levels.
We release \bench, 10K verifiable image generation tasks where the model is asked to draw geometric objects with specified color, shape, count, and spatial relations, and show that models struggle with visual instruction following. 
A \bench-Challenge set where the strongest frontier image generation model, GPT-Image-2.5-Sunburst, solves only 21.4\% of the tasks.
We then train image generators with VVR scores as an RL reward (RLVVR) significantly improves instruction following both on VVR tasks and on natural prompts unseen during training. 
Mixing VVR with existing post-training objectives for image generation, such as GenEval2, leads to further gains on a broad evaluation suite and human preference, motivating its adoption into standard post-training recipes.

%% file: sec/7_limitations.tex
\section*{Limitations and future directions}
\label{sec:limitations}

\bench uses eight colors, three shapes, and plain backgrounds; future work can add more shapes, textures, and object types as new program verifiers.
VVR currently covers 2D geometric objects, and future work can extend it to 3D renderings or 2D projections of 3D objects.

VVR is constrained to text-to-image generation; the same constraints could be applied to image editing. New constraints such as motion, velocity, acceleration, are also convertible to program verifiers and can be applied to video generation.
VVR outputs with their verifier decisions could be used to evaluate or train learned reward models and VLM judges.

We post-train SD3.5-M with Flow-GRPO; applying RLVVR to other image generators and RL algorithms is left to future work.
RLVVR is an RL-Zero recipe: we apply Flow-GRPO directly to the pretrained SD3.5-M.
Mid-training on VVR data with supervised fine-tuning or DPO before the RL stage, with different data mixtures, could further improve instruction following in image generation.

RLVVR uses the combined dense reward $r_{\mathrm{dense}}$, but the program verifiers also report which constraints fail.
This feedback allows a range of reward designs, for example weighting constraint families differently according to the desired model behavior.
The complexity-controlled task generator also allows adaptive curricula for RLVVR.

Several API models incorrectly decline some VVR tasks as contradictory, although every task is satisfiable. Our analysis is constrained to case studies due to the small number of abstentions, but VVR tasks can be used to evaluate, and further train for, correct abstention decisions in image generators, VLM, and even LLMs to improve spatial reasoning.

%% file: sec/7_statements.tex
\subsection*{AI use statement}

Generative AI tools were used to assist with code navigation, debugging, analysis scripting, and manuscript polishing.
The authors take responsibility for the final content.

\subsection*{Ethics statement}

The annotation in this paper labels generated images of synthetic scenes and public benchmark prompts and involves no personal or sensitive data.

\subsection*{Reproducibility statement}

We release all three benchmark splits, the two training corpora, the verifier, the task generator, and the scripts that build every table and figure, together with evaluation prompts, training configurations, and model checkpoints.
The appendix records reward formulas, full results tables, and dataset statistics.

\section*{Acknowledgment}
This research was developed in part with funding from the Defense Advanced Research Projects Agency's (DARPA) SciFy program (Agreement No. HR00112520300). The views expressed are those of the author and do not reflect the official policy or position of the Department of Defense or the U.S.~Government. 
This material is based in part upon work supported by the Defense Advanced Research Projects Agency and the Air Force Research Laboratory, contract number(s): FA8650-23-C-7316. Any opinions, findings and conclusions, or recommendations expressed in this material are those of the author(s) and do not necessarily reflect the views of AFRL or DARPA.
This research was supported by Coefficient Giving, the University of Washington Population Health Initiative, Amazon Health, the UW+Amazon Science Hub, and the Meta AIM program.

%% file: app/appendix.tex
\newpage
\appendix

\section{Task Representation}
\label{app:representation}

This section lists the constraint library of \S\ref{sec:representation} and gives a complete example task.

\subsection{Constraint library}
\label{app:constraint-library}

Table~\ref{tab:vvr-verifier-bank} lists all 46 constraint types and their contributions to structural complexity.
Let $L(n)=1+\log_2 n$; $n_i$ is the number of visible instances in object group $i$, $N_l$ is the number of objects checked by layout constraint $l$, $N_r$ is the total number of visible instances in the distinct groups referenced by relation $r$, and $m_r$ is the number of required one-to-one matches.

\input{tables/vvr_bench_verifier_bank}

\subsection{Example task}
\label{sec:representation-example}

This example gives one task in the representation of \S\ref{sec:representation} and its prompt; Appendix~\ref{app:program-verifiers} gives the verifier code for each of its constraints.

\paragraph{Task.}
The task contains two object groups and seven constraints:
\begin{align*}
\mathcal{G}=\{&\texttt{g1},\texttt{g2}\},\\
\mathcal{A}=\{&\texttt{color\_attribute(g1;\,purple)},\ \texttt{shape\_attribute(g1;\,circle)},\ \texttt{exact\_count(g1;\,1)},\\
=\{&\texttt{color\_attribute(g2;\,yellow)},\ \texttt{shape\_attribute(g2;\,square)},\ \texttt{exact\_count(g2;\,1)},\\
&\texttt{below(g1,g2)}\}.
\end{align*}
The background constraints are $\mathcal{B}=\{\texttt{background\_color(pale pink)}\}$, and the forbidden-content constraints are $\mathcal{F}=\{\texttt{no\_unrequested\_objects}\}$.
For compactness, the implementation stores the color, shape, and count constraints of each group within the group's record, stores the background constraint as the background color, and applies \texttt{no\_unrequested\_objects} to every task, so its \texttt{forbidden} list holds only additional forbidden-content constraints:
\begin{verbatim}
{
  "background": {"color": "pale pink"},
  "objects": [
    {"id": "g1", "color": "purple", "shape": "circle", "count": 1},
    {"id": "g2", "color": "yellow", "shape": "square", "count": 1}
  ],
  "relations": [{"type": "below", "subject": "g1", "object": "g2"}],
  "forbidden": []
}
\end{verbatim}
A group with \texttt{"count\_mode": "relative"} has no \texttt{exact\_count} constraint; its count is constrained only by relations such as \texttt{times\_as\_many}.

\paragraph{Prompt.}
The templates produce ``Place the purple circle below the yellow square.
Set the objects against a plain pale pink background; do not add other colored objects.''
Each constraint refers to groups by identifier, so the binding of each attribute to its object is unambiguous.

\section{Task Generation}
\label{app:vvr_generator}
\label{sec:generation-details}

This section gives the complete generation procedure of \S\ref{sec:task-generation}, its validation steps, a worked example, and the construction of each dataset.

\subsection{Procedure}

Algorithm~\ref{alg:vvr-sampling-full} gives the procedure for one dataset.
Its first steps implement steps 1--5 of \S\ref{sec:task-generation}, and the remaining steps are the validation checks of Appendix~\ref{app:generation-validation}.

\begin{algorithm}[H]
\caption{Task generation and validation for one dataset.}
\label{alg:vvr-sampling-full}
\begin{algorithmic}[1]
\State initialize an empty task pool $\mathcal{Q}$
\For{each constructor index}
  \State sample background constraints $\mathcal{B}$, object groups $\mathcal{G}$ with colors, shapes, and counts, and forbidden-content constraints $\mathcal{F}$
  \State assign each object instance a center and a size to obtain the scene $z$
  \State $\mathcal{A}^*\gets$ the instantiated constraints whose requirements hold in $z$
  \For{each active constraint set $\mathcal{A}\subseteq\mathcal{A}^*$ selected within the target complexity range}
    \State render the prompt $p$ from $(\mathcal{G},\mathcal{B},\mathcal{A},\mathcal{F})$ and form $s=(\mathcal{G},\mathcal{B},\mathcal{A},\mathcal{F},p)$
    \State render the reference image $x^\star$ of $z$ on the background specified by $\mathcal{B}$
    \If{$p$ or the canonical form of $s$ occurs in $\mathcal{Q}$ or in an excluded split}
      \State \textbf{continue}
    \EndIf
    \If{$r_{\mathrm{exact}}(x^\star,s)=0$}
      \State \textbf{continue}
    \EndIf
    \State change one constraint of $s$ to obtain a counterfactual task $\tilde{s}$
    \If{the changed constraint passes on $x^\star$ under $\tilde{s}$}
      \State \textbf{continue}
    \EndIf
    \State add $(s,x^\star)$ to $\mathcal{Q}$
  \EndFor
\EndFor
\State select tasks from $\mathcal{Q}$ to match the target distribution of the dataset (Appendix~\ref{app:generation-datasets})
\end{algorithmic}
\end{algorithm}

\paragraph{Scene construction.}
The frozen constructors represent each object instance by its group identifier, integer center $(u,v)$, and radius $\rho$ on a $512\times512$ canvas.
The default layout divides the canvas into six boxes arranged in three columns and two rows.
For a group of $n$ repeated objects, the constructor uses $\min(5,\lceil\sqrt{n}\rceil)$ columns and fills $\lceil n/\min(5,\lceil\sqrt{n}\rceil)\rceil$ rows at evenly spaced coordinates within its box.
Relation-specific templates replace these default placements with fixed constructions for rows, columns, grids, contact, containment, order, proximity, extrema, and relative size.
Seeded sampling selects counts, attributes, and template variants.
The constructor then enumerates additional constraints that are true of the stored positions and sizes.

\subsection{Validation}
\label{app:generation-validation}

Every retained task passes three checks.
\paragraph{Reference image.}
The generator renders the scene and requires the released verifier to accept it, which confirms that the pixel rendering preserves every constraint that holds in the scene.
\paragraph{Counterfactual.}
The generator changes one constraint while holding the reference image fixed and requires the verifier to reject the image under the changed task.
For \bench, the changed constraint is evaluated on its own; for Challenge and the scene-first training candidates, the generator inverts one constraint and removes the other relation and layout constraints that could conflict with the inversion.
\paragraph{Deduplication.}
Deduplication uses normalized prompts and canonical tasks formed by renaming object identifiers in a fixed order, and it is applied jointly across each dataset and all excluded training and evaluation splits.

For each of the 10,000 \bench and 720 Challenge tasks, the verifier accepts the reference image and rejects the counterfactual.

\subsection{Worked example}

This example traces one scene through the five steps of \S\ref{sec:task-generation}.
It is the output of the scene-first generator for enumeration index 53 with the default seed; every value below is produced by the code.

\paragraph{Step 1: scene.}
The generator samples $\mathcal{B}=\{\texttt{background\_color(white)}\}$ and three object groups with seven objects in total on a $512\times512$ canvas; $\mathcal{F}=\{\texttt{no\_unrequested\_objects}\}$.
\begin{center}
\small
\begin{tabular}{@{}lllrl@{}}
\toprule
Group & Color & Shape & Count & Centers (radius 9 px) \\
\midrule
\texttt{g0} & cyan & circle & 3 & $(28,34)$, $(148,34)$, $(28,214)$ \\
\texttt{g1} & yellow & square & 2 & $(196,34)$, $(316,34)$ \\
\texttt{g2} & pink & triangle & 2 & $(364,34)$, $(484,34)$ \\
\bottomrule
\end{tabular}
\end{center}

\paragraph{Steps 2 and 3: satisfiable constraint set.}
The generator instantiates each constraint type on the group tuples of its arity and keeps the instantiated constraints that hold in the scene.
The resulting set $\mathcal{A}^*$ contains the nine unary color, shape, and count constraints of the three groups and the following fifteen constraints:
\begin{itemize}[itemsep=0pt,topsep=2pt,leftmargin=15pt]
    \item count comparisons: \texttt{more\_than\_count(g0,g1)}, \texttt{more\_than\_count(g0,g2)}, \texttt{same\_count(g1,g2)};
    \item order: \texttt{all\_left\_of(g0,g1)}, \texttt{all\_left\_of(g0,g2)}, \texttt{all\_left\_of(g1,g2)};
    \item alignment: \texttt{not\_all\_same\_row(g0)}, \texttt{not\_all\_same\_column(g0)}, \texttt{all\_same\_row(g1)}, \texttt{not\_all\_same\_column(g1)}, \texttt{all\_same\_row(g2)}, \texttt{not\_all\_same\_column(g2)};
    \item regions: \texttt{absolute\_region(g0;\,top)}, \texttt{absolute\_region(g1;\,top)}, \texttt{absolute\_region(g2;\,top)}.
\end{itemize}
For each pair of groups, the generator adds the one count comparison that holds and a direction only when the groups are separated by at least 16 px along that axis, a margin wider than the verifier's 12 px.

\paragraph{Step 4: active constraints.}
The generator adds at most four constraints from $\mathcal{A}^*$ to the unary ones, visiting constraint types in a fixed rotated order and skipping any constraint that would exceed the target complexity range.
Every prefix of this sequence whose complexity lies in the target range is a task, so this scene yields four nested tasks.
The largest adds \texttt{all\_same\_row(g2)}, \texttt{not\_all\_same\_row(g0)}, \texttt{not\_all\_same\_column(g2)}, and \texttt{same\_count(g1,g2)}.
Because \texttt{same\_count} fixes the number of yellow squares relative to the pink triangles, \texttt{g1} loses its \texttt{exact\_count} constraint, so $\mathcal{A}$ contains twelve constraints: color and shape for all three groups, \texttt{exact\_count(g0;\,3)}, \texttt{exact\_count(g2;\,2)}, and the four added constraints.
Its structural complexity is
\begin{equation*}
C(s)=\underbrace{3L(3)}_{\texttt{g0}}+\underbrace{2L(2)}_{\texttt{g1}}+\underbrace{3L(2)}_{\texttt{g2}}+\underbrace{L(2)+L(3)+L(2)+L(4)}_{\text{added constraints}}=27.34,
\end{equation*}
with $L(n)=1+\log_2 n$.

\paragraph{Step 5: prompt.}
The templates render the four nested tasks with different sentence frames:
\begin{itemize}[itemsep=0pt,topsep=2pt,leftmargin=15pt]
    \item ``The image should contain three cyan circles, two yellow squares, and two pink triangles. Arrange all the pink triangles in one row. Use a plain white background and no other colored objects.''
    \item ``Show three cyan circles, two yellow squares, and two pink triangles. Arrange all the pink triangles in one row. Arrange all the cyan circles so they are not all in the same row. Keep the background plain white, with no additional colored objects.''
    \item ``Create an image with three cyan circles, two yellow squares, and two pink triangles. Arrange all the pink triangles in one row. Arrange all the cyan circles so they are not all in the same row. Arrange all the pink triangles so they are not all in the same column. Set the objects against a plain white background; do not add other colored objects.''
    \item ``Draw three cyan circles and two pink triangles. Use the same number of yellow squares and pink triangles. Arrange all the pink triangles in one row. Arrange all the cyan circles so they are not all in the same row. Arrange all the pink triangles so they are not all in the same column. Use a plain white background and no other colored objects.''
\end{itemize}

\begin{wrapfigure}{r}{0.24\textwidth}\vspace{-6mm}
\centering
\includegraphics[width=\linewidth]{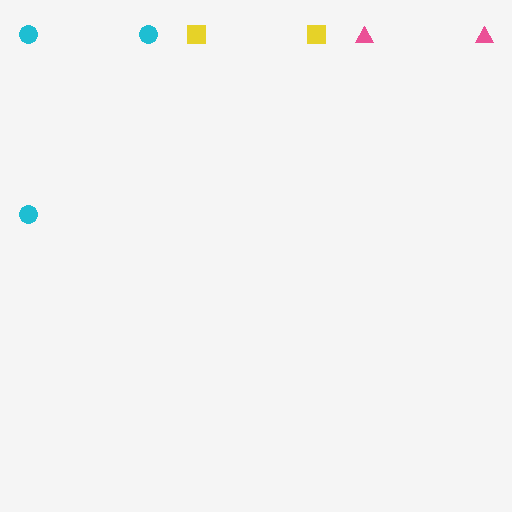}\vspace{-7mm}
\caption{Reference image of the example scene.}
\label{fig:app-generator-example}\vspace{-14mm}
\end{wrapfigure}

The last prompt states no count for the yellow squares, matching the removal of their \texttt{exact\_count} constraint.

\paragraph{Validation.}
The generator renders the scene as the reference image in Figure~\ref{fig:app-generator-example}, and the released verifier accepts it for all four tasks.
The counterfactual of each task replaces \texttt{all\_same\_row(g2)} with \texttt{not\_all\_same\_row(g2)} and removes the other added constraints; the verifier rejects the same image under every counterfactual.

\subsection{Structural complexity}

With $L(n)=1+\log_2 n$, the complexity of a task is
\begin{equation}
\label{eq:vvr-complexity}
C(s)=\sum_{a\in\mathcal{A}}c(a;s),
\end{equation}
where the cost $c(a;s)$ of each constraint type is given in Table~\ref{tab:vvr-verifier-bank}.
Color, shape, and exact count cost $L(n)$ for a group of $n$ objects; most relations and layouts cost $L(N)$ for the $N$ objects they compare; a count ratio by factor $k$ adds $\log_2 k$; one-to-one containment costs the number of required matches; and within-group size variation costs $2\log_2 N$.
Every task has one background-color constraint and the forbidden-content constraint \texttt{no\_unrequested\_objects}, so the constraints in $\mathcal{B}$ and $\mathcal{F}$ are excluded.

\paragraph{Complexity as a predictor of failure.}
\label{sec:complexity-validation}
For each model we compute the AUC with which a single task feature separates unsolved from solved \bench tasks, and the McFadden $R^2$ of a logistic regression of exact success on that feature.
The features are $C(s)$, the number of color, shape, relation, and layout constraints, and the numbers of object instances, object groups, and relations.
The 19 models are the ten models of Table~\ref{tab:vvr-bench-foundation-models} other than Sana and SDXL, which solve fewer than three tasks, and the nine post-trained SD3.5-M models of Table~\ref{tab:vvr-bench-sd35-posttraining}.
$C(s)$ has the highest AUC and $R^2$ for 18 models, with median AUC 0.857 and $R^2$ 0.283; the number of color, shape, relation, and layout constraints follows with 0.829 and 0.232, and the number of object instances with 0.820 and 0.227.

\subsection{Datasets}
\label{app:generation-datasets}

Each dataset is built by generating a pool of validated candidates with Algorithm~\ref{alg:vvr-sampling-full} and selecting tasks from the pool to match a target distribution (Table~\ref{tab:dataset-targets}).
All datasets use eight foreground colors, three shapes, counts from one through ten, and seven backgrounds, with reference images on a $512\times512$ canvas, and no selection step uses model outputs.
Candidates are organized into nine generation strata: quantity, binding, location, direction and order, between, proximity, size, structured layout, and topology.

\begin{table}[H]
\centering
\small
\caption{Candidate pool and target distribution of each dataset.}
\label{tab:dataset-targets}
\begin{tabular}{@{}lr>{\raggedright\arraybackslash}p{0.28\textwidth}>{\raggedright\arraybackslash}p{0.32\textwidth}@{}}
\toprule
Dataset & Size & Candidates & Target distribution \\
\midrule
\bench & 10,000 & Single strata and compositions of two to six strata at five scene-size settings & Complexity 3--48, capped at each integer complexity \\
\bench-Fast & 820 & \bench & 20 tasks at each attainable integer complexity from 3 to 44 \\
\bench-Challenge & 720 & Scenes seeded by each of the 46 constraint types, with up to six added relation or layout constraints & 20 tasks at each integer complexity from 45 to 80, and at least 20 tasks per non-grounding constraint type \\
VVR-Easy & 100,000 & One constraint type from one stratum & Complexity at most 20, equal quotas over the nine strata \\
VVR-Matched & 100,000 & The \bench and \bench-Challenge generators & The strata and complexity distribution of \bench \\
\bottomrule
\end{tabular}
\end{table}

\paragraph{\bench.}
The generator crosses each stratum and each composition of two, three, and four to six strata with five scene-size settings, which control the number of object groups and instances.
Each single stratum receives 50 candidates per setting, and each composition order receives 600 candidates per setting, divided evenly over stratum combinations, for 11,250 candidates.
Selection caps the number of tasks at each integer complexity by removing candidates from the most populated complexities, and adds single-object tasks at complexities the generator does not otherwise reach.
\bench uses 32 of the 46 constraint types; the remaining 14 appear only in \bench-Challenge (Table~\ref{tab:vvr-verifier-bank}).

\paragraph{\bench-Challenge.}
The first constraint of each candidate cycles through all 46 constraint types, and its reference image is constructed to satisfy it.
The generator then adds at most one constraint of each type, skipping duplicate relations and combinations that cannot hold together, and every prefix of the added constraints is a candidate.
Besides the targets in Table~\ref{tab:dataset-targets}, selection allows at most two thirds of a task's complexity to come from color, shape, and exact count constraints, and balances complexity across families and constraint types within each family.

\paragraph{VVR-Easy.}
Each task adds one constraint type to the color, shape, and count constraints of its object groups, so it exercises at most one constraint family beyond them.
Constraint types within each stratum receive fixed quotas.

\paragraph{VVR-Matched.}
Tasks are allocated to strata in proportion to \bench, and family and constraint-type frequencies are equalized within each stratum.
The corpus is accepted only if a Kolmogorov--Smirnov test finds its complexity distribution matched to that of \bench.

\section{Verifier}
\label{sec:verifier-semantics}

This section describes the verifier of \S\ref{sec:verification}: object extraction, the program verifiers, the scores and training reward, and its validation.

\subsection{Pixel-to-object extraction}
\label{sec:pixel-object-details}

The verifier estimates the background as the median color along the image boundary and uses variation among those boundary pixels to set a background-relative foreground threshold.
It converts the image to HSV and assigns sufficiently saturated foreground pixels to fixed, nonoverlapping hue ranges for the eight supported colors.
Low-confidence and background-like pixels are excluded.
On each binary color mask, erosion followed by dilation removes isolated foreground pixels, and dilation followed by erosion fills small holes and narrow breaks.
The implementation scans the cleaned mask and uses flood fill from each unlabeled foreground pixel, traversing horizontal, vertical, and diagonal neighbors.
Every maximal set reached by one traversal becomes a candidate object.

\begin{figure}[t]
\centering
\includegraphics[width=\linewidth]{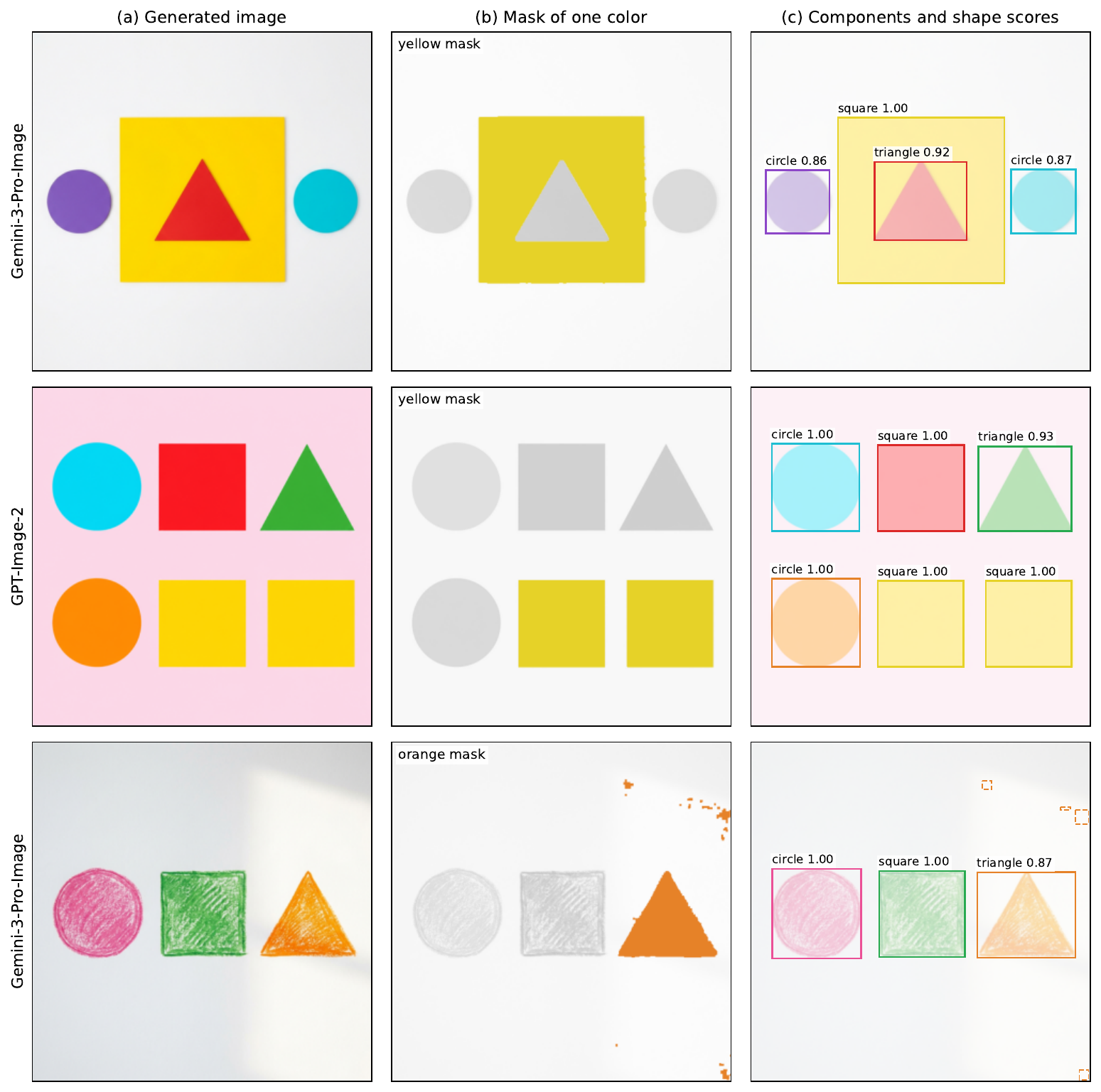}
\caption{Object extraction on three \bench-Fast outputs of API models.
(a) The generated image.
(b) The cleaned mask of one requested color over the grayed-out image.
(c) The connected components of all requested colors, each labeled with the shape that receives its highest score.
The bottom image is a textured crayon drawing under uneven light: the orange mask still covers the whole triangle, and the lit background adds small orange fragments (dashed boxes) that are too small to count as objects.
The verifier accepts all three images.
Prompts: (top) ``Place the red triangle inside the yellow square. Add a purple circle and a cyan circle as well. Keep the background plain white, with no additional colored objects.''
(middle) ``Show a cyan circle, a red square, a green triangle, an orange circle, and two yellow squares. Use a plain pale pink background and no other colored objects.''
(bottom) ``The image should contain a pink circle, a green square, and an orange triangle. Set the objects against a plain white background; do not add other colored objects.''}
\label{fig:verifier-extraction}
\end{figure}

\begin{figure}[t]
\centering
\includegraphics[width=\linewidth]{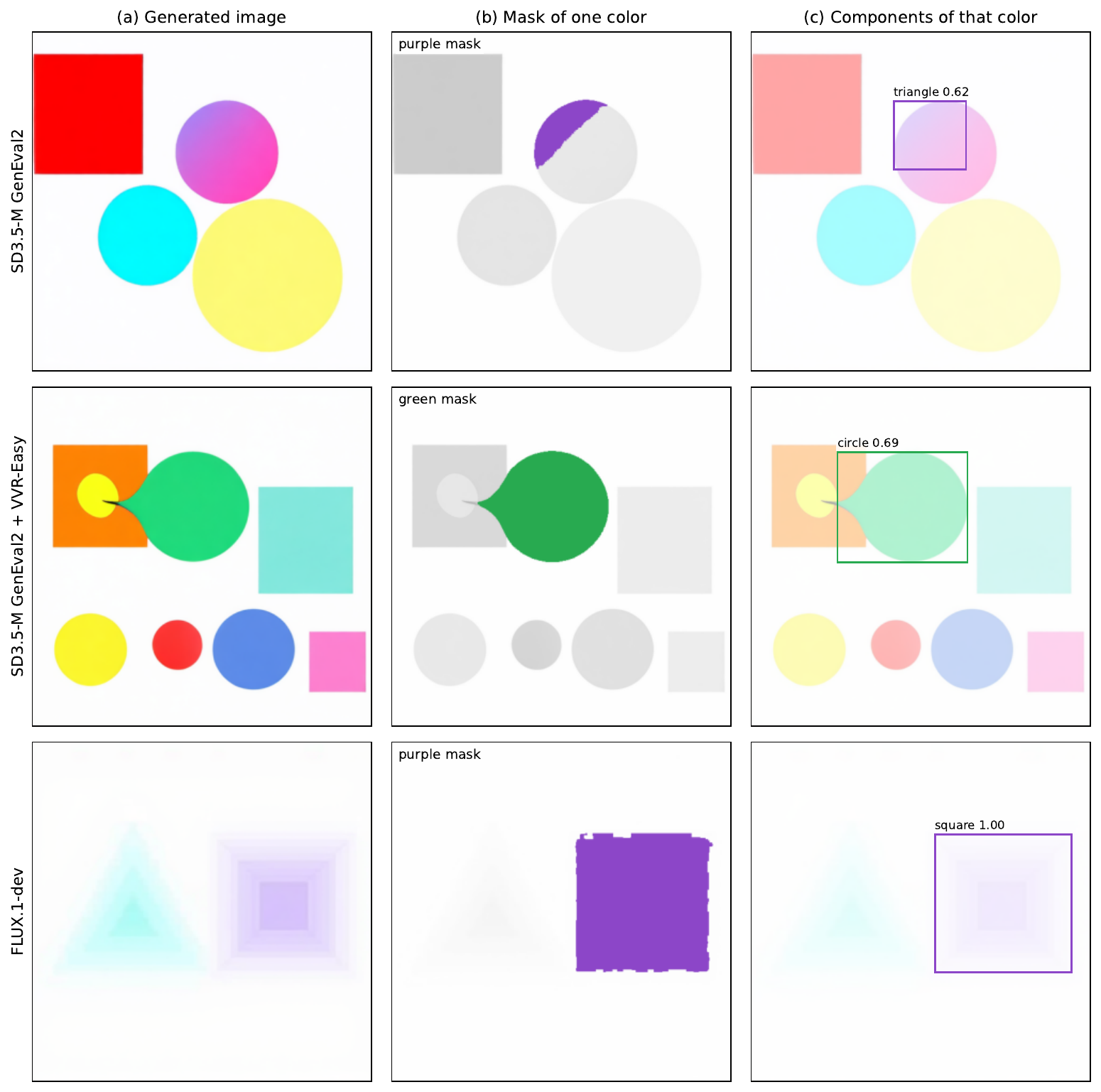}
\caption{Object extraction on distorted generations of open-weight models.
Panels as in Figure~\ref{fig:verifier-extraction}, except that (c) shows only the components of the highlighted color.
(Top, ambiguous color) The circle requested as purple shades from purple into pink; the purple mask covers only its upper part, whose highest shape score is triangle 0.62.
(Middle, irregular contour) The green circle grows a tail that reaches into the orange square; its circle score drops to 0.69, compared with 0.89 to 0.99 for the undistorted circles in the same image.
(Bottom, blurred boundaries) The purple mask covers the whole blurred square, which scores 1.00 as a square; the image still fails because the prompt asks for two cyan triangles and two purple squares and the image shows one of each.
The verifier rejects all three images.}
\label{fig:verifier-extraction-distorted}
\end{figure}

Each component is described by its area, centroid, bounding box, boundary, aspect ratio, bounding-box occupancy, convexity, convex-hull vertex count, number of holes, and offset between its centroid and bounding-box center.
A fixed geometric classifier converts these measurements into circle, square, and triangle scores.
Circle scores favor approximately equal width and height, high convexity, and rounded contours; square scores cover both filled axis-aligned boxes and centered, convex rotated squares; triangle scores use their characteristic bounding-box occupancy and off-center centroid.
A single hole provides additional evidence for an outlined circle or square.
Components that are too small, narrow, or weakly supported by the requested color and shape are removed.
If several requested groups have the same color but different shapes, each component is assigned exclusively to the shape receiving its highest score.
Figure~\ref{fig:verifier-extraction} shows the color masks, components, and shape scores for three API model outputs, and Figure~\ref{fig:verifier-extraction-distorted} shows the same steps on distorted open-weight generations with ambiguous colors, irregular contours, and blurred boundaries.

The frozen rules include three safeguards for imperfect generations.
First, background-adaptive contrast and calibrated hue boundaries handle shading and colors near category boundaries.
Explicit boundary rules separate pale, low-saturation red from pink and muted blue-violet from bright blue.
Second, morphological cleanup and a shape-conditioned fallback mask recover objects with fragmented or blurred color regions without allowing one component to satisfy two color groups.
Third, robust extents use the 5th and 95th percentiles of component coordinates, reducing sensitivity to stray boundary pixels.
The exact thresholds are fixed in the released verifier.
Appendix~\ref{sec:verifier-validation-details} reports calibration and held-out tests covering ambiguous colors, irregular contours, compression, blur, touching objects, and threshold-adjacent cases.

\subsection{Program verifiers}
\label{app:program-verifiers}

The listings below are excerpts from the released \texttt{vvr\_bench/verifier.py} for the four constraint types in the example task of Appendix~\ref{sec:representation-example}.
Helper functions are named but not shown.
The extraction step of \S\ref{sec:verification} provides each group's matched objects as components with a centroid and a score for each shape; \texttt{\_estimate\_repeated\_group\_count} counts the objects of a group and counts a connected region whose area is close to an integer multiple of one object's area as that many touching objects.

For \texttt{exact\_count} and \texttt{color\_attribute}, the verifier compares the estimated count with the target and requires at least one object of the group's color:
\begin{lstlisting}[language=Python,basicstyle=\ttfamily\scriptsize,frame=single,framesep=4pt,columns=fullflexible,showstringspaces=false]
count_pred = _estimate_repeated_group_count(count_components, shape)
if count_is_exact:                      # exact_count
    count_error, count_score = _score_exact_count(count_pred, target_count)
else:                                   # count_mode == "relative": presence only
    count_error = 0.0 if count_pred >= 1 else 1.0
    count_score = 1.0 if count_pred >= 1 else 0.0

color_presence_score = min(1.0, float(count_pred))   # color_attribute
color_presence_strict = count_pred >= 1
if (len(spec.get("objects", [])) > 1 and not color_presence_strict
        and any(_component_identity_compatible(c, shape, image.shape[:2])
                for c in fallback_components)):
    color_presence_score = 1.0
    color_presence_strict = True

def _score_exact_count(observed_count, target_count):
    error = abs(observed_count - target_count)
    score = max(0.0, 1.0 - error / max(target_count, 1))
    return error, float(score)
\end{lstlisting}
The \texttt{exact\_count} constraint passes when \texttt{count\_error} is zero, and \texttt{color\_attribute} passes when \texttt{color\_presence\_strict} holds.

For \texttt{shape\_attribute}, the verifier averages the requested shape's score over the group's objects and compares it with a shape-specific threshold:
\begin{lstlisting}[language=Python,basicstyle=\ttfamily\scriptsize,frame=single,framesep=4pt,columns=fullflexible,showstringspaces=false]
shape_score = _score_shape_attribute(shape, selected, allow_occluded_triangle=True)
shape_strict = shape_score >= _shape_presence_threshold(shape)

def _score_shape_attribute(shape, components, *, allow_occluded_triangle=False):
    if not components:
        return 0.0
    return float(np.mean([
        _effective_shape_score(component, shape) if allow_occluded_triangle
        else component.shape_scores.get(shape, 0.0)
        for component in components
    ]))

def _shape_presence_threshold(shape):
    return 0.20 if shape == "triangle" else 0.40
\end{lstlisting}

For \texttt{below}, the verifier first rejects nested referents, where one group's object lies inside the other's, and then compares the mean vertical centroids with a margin; image coordinates increase downward:
\begin{lstlisting}[language=Python,basicstyle=\ttfamily\scriptsize,frame=single,framesep=4pt,columns=fullflexible,showstringspaces=false]
nested = any(
    _bbox_intersection_fraction(first, second) >= 0.98
    and math.dist(first.centroid, second.centroid)
        <= 0.80 * min(_component_extent(first), _component_extent(second))
    and max(first.shape_scores.values(), default=0.0) >= 0.40
    and max(second.shape_scores.values(), default=0.0) >= 0.40
    for first in subject
    for second in obj
)
if nested:
    return 0.0, {**result, "strict_pass": False, "nested_referents": True}

subj_y = float(np.mean([comp.centroid[1] for comp in subject]))
obj_y = float(np.mean([comp.centroid[1] for comp in obj]))
margin = float(relation.get("margin_px", 24.0))
delta = subj_y - obj_y
score = min(1.0, max(0.0, delta / max(margin, 1.0)))
strict_pass = delta >= margin
\end{lstlisting}
The verifier also reports a color--shape binding score for each group, computed from its color and shape scores; binding adds no structural complexity (Table~\ref{tab:vvr-verifier-bank}).
The image passes this task when all seven constraints in $\mathcal{A}$ and the constraints in $\mathcal{B}$ and $\mathcal{F}$ pass (\S\ref{sec:verification}).

\paragraph{Constraint measurements.}
The verifier assigns a fixed geometric meaning to each relational phrase in the prompt templates.
Let $h$ be the shorter image side and $e$ the largest visible extent among the objects compared.
\begin{itemize}[itemsep=0pt,topsep=2pt,leftmargin=15pt]
\item \emph{Same row} (column): the vertical (horizontal) spread of the object centroids is at most $\max(0.04h,\,0.55e)$.
\item \emph{Between}: let $t$ be the position of the subject's centroid projected onto the segment joining the two reference centroids, and $d$ its distance from that segment.
The relation holds when $\max(0,1-d/0.25\ell)\cdot\mathbf{1}[0.15\le t\le0.85]\ge0.70$, where $\ell$ is the segment length and the indicator is replaced by a linear decay outside the interval for partial credit.
\item \emph{Closer than}: distance is the minimum Euclidean distance between component boundaries, which reflects the visible gap between objects of different sizes.
The nearer distance must be at most 0.90 of the farther distance and at least 4 pixels smaller.
\item \emph{Largest} (smallest) \emph{colored object}: the subject's visual extent, defined below, is compared with that of every visible colored component in the image, including components that match no requested group.
\end{itemize}
Relative size uses visual extent, the geometric mean of a component's width and height measured between the 5th and 95th percentiles of its pixel coordinates.
The benchmark compares sizes only relative to other objects, because calibration found no stable human decision boundary for absolute size.

\paragraph{Objects and unmatched components.}
Requested objects are matched by color and shape to connected visual components.
Count compares the number of matched components with the requested cardinality.
Any remaining visible colored component is unmatched, so an extra copy of a requested object lowers both the count score and the unmatched-component score.

\subsection{Verifier validation}
\label{sec:verifier-validation-details}

The verifier passes 290 historical edge cases, 360 direct checks, 14,788 metamorphic checks, a 320-case matrix of single constraints, 200 constructed cases at decision thresholds, and geometry tests for repeated-group size, containment, contact, and relation inverses.
These tests cover blur, compression, low contrast, irregular contours, touching and merged components, missing objects, and reversed relations.
During development, 3,947 human decisions set the decision boundary of each perceptual predicate: the hue range of every color name, the margin at which two objects touch, the contour tolerances that separate circles, squares, and triangles, and the ratio at which one object counts as larger than another.

Two human audits test the verifier on generated images.
Each audit image tests one constraint, labeled by one annotator without seeing the verifier's decision.
The larger audit contains 512 images of 128 prompts generated by pretrained SD3.5-M, FLUX.1-dev, and two SD3.5-M models trained with earlier VVR rewards.
The verifier version frozen before this audit agrees with 454 of 508 decisive labels (89.4\%, Cohen's $\kappa=0.78$), and an earlier audit of 528 images agrees on 421 of 452 (93.1\%, $\kappa=0.86$).
After calibration that used the larger audit, the released verifier, which scores every result in this paper, agrees with 481 of its 508 labels (94.7\%, $\kappa=0.89$).

\subsection{Scores and training reward}
\label{app:scores}

The dense reward $r_{\mathrm{dense}}$ of Eq.~\ref{eq:dense_reward} combines the partial-credit scores $q_a$ in two levels.
Each requested object group $i$ receives
\begin{equation}
 r_i=0.45r_{\mathrm{count},i}+0.20r_{\mathrm{shape},i}
     +0.20r_{\mathrm{layout},i}+0.15r_{\mathrm{size},i},
\end{equation}
where each term is the partial-credit score of that group's constraints of the given kind, and $r_{\mathrm{obj}}$ is the mean across groups.
Let $r_{\mathrm{rel}}$ be the mean partial-credit score of the relations, $I_{\mathrm{rel}}$ indicate whether the task has a relation, and $r_{\mathrm{forbid}}$, $r_{\mathrm{extra}}$, and $r_{\mathrm{bg}}$ be the scores of the forbidden-content, unmatched-component, and background constraints.
The weighted sum in Eq.~\ref{eq:dense_reward} is
\begin{equation}
 \sum_{a}w_aq_a =
 \frac{0.55r_{\mathrm{obj}}+0.15I_{\mathrm{rel}}r_{\mathrm{rel}}+
 0.15r_{\mathrm{forbid}}+0.10r_{\mathrm{extra}}+
 0.05r_{\mathrm{bg}}}{0.85+0.15I_{\mathrm{rel}}},
\end{equation}
and the penalty factor is
\begin{equation}
\label{eq:vvr-training-reward}
 \psi = g_{\mathrm{count}}g_{\mathrm{rel}}g_{\mathrm{extra}}g_{\mathrm{forbid}},
\end{equation}
with
\begin{equation}
\begin{split}
g_{\mathrm{count}} &= 0.10+0.90r_{\mathrm{count}},\\
g_{\mathrm{rel}} &=
\begin{cases}
1, & I_{\mathrm{rel}}=0,\\
0.25+0.75r_{\mathrm{rel}}, & I_{\mathrm{rel}}=1,
\end{cases}\\
g_{\mathrm{extra}} &= \max\!\left(0,1-\frac{n_{\mathrm{extra}}}
{\max(N_{\mathrm{target}},1)}\right),\\
g_{\mathrm{forbid}} &= r_{\mathrm{forbid}}.
\end{split}
\end{equation}
Here $r_{\mathrm{count}}$ is the mean group-level count score, $n_{\mathrm{extra}}$ is the number of unmatched components, and $N_{\mathrm{target}}$ is the requested object count.

\section{Benchmark Evaluation Details}

\subsection{Evaluation details}
\label{sec:evaluation-details}

Models generate at their native resolution.
We score \bench images at $512\times512$ and Challenge images at $1024\times1024$, which preserves boundaries and small objects in dense scenes.
API models receive one request per prompt; transient errors are retried, and completed responses are never resampled.
Appendix~\ref{sec:no-image-responses} reports how often each API model returned no image.

\paragraph{Generation settings.}
Table~\ref{tab:generation-settings} lists the settings of every model.
Open-weight models generate at $1024\times1024$, except HunyuanImage-2.1 at $2048\times2048$, and all post-trained SD3.5-M models use the SD3.5-M settings.
The seed of each prompt is the first 32 bits of the SHA-256 hash of a fixed base seed and the prompt identifier, so every open-weight model receives the same seed for the same prompt.
The OpenAI image API has no temperature parameter, and Gemini models are called with a 1:1 aspect ratio and default values for temperature and all other sampling parameters.

\begin{table}[H]
\centering
\scriptsize
\setlength{\tabcolsep}{4pt}
\caption{Generation settings. Guidance is the classifier-free guidance scale.}
\label{tab:generation-settings}
\begin{tabular}{@{}lrr@{\hspace{12pt}}ll@{}}
\toprule
Open-weight model & Steps & Guidance & API model & Settings \\
\midrule
FLUX.2-dev & 50 & 4.0 & GPT-Image-2.5-Sunburst (2026-09-08) & medium quality, $1024\times1024$ \\
HunyuanImage-2.1 & 50 & 3.5 & GPT-Image-2 (2026-04-21) & medium quality, $1024\times1024$ \\
Qwen-Image-2512 & 50 & 4.0 & GPT-Image-1-mini & medium quality, $1024\times1024$ \\
HiDream-I1-Full & 50 & 5.0 & Gemini-3-Pro-Image & 1K \\
FLUX.1-dev & 28 & 3.5 & Gemini-3.1-Flash-Image & 1K \\
FLUX.1-schnell & 4 & 0.0 & Gemini-3.1-Flash-Lite-Image & 1K \\
SD3.5 Medium & 40 & 4.5 & Gemini-2.5-Flash-Image & model default (1K) \\
SD3.5 Large & 40 & 4.5 & & \\
SDXL 1.0 & 40 & 5.0 & & \\
Sana 1.6B & 20 & 4.5 & & \\
\bottomrule
\end{tabular}
\end{table}

\subsection{Complete \bench-Fast results}
\label{sec:vvr-core820-details}

Table~\ref{tab:vvr-bench-core-api-820} reports the complete complexity breakdown underlying Figure~\ref{fig:vvr-bench-core-api-820}.

\input{tables/vvr_bench_core_api_820}

\subsection{Responses without an image}
\label{sec:no-image-responses}

Some API models return text instead of an image, typically stating that the prompt is contradictory or too complex.
These abstentions are incorrect: every task is satisfiable, because its reference image passes the verifier.
On Challenge, Gemini-2.5-Flash-Image returned no image for 39 of 720 prompts, Gemini-3.1-Flash-Image for 3, Gemini-3.1-Flash-Lite-Image for 2, and Gemini-3-Pro-Image for 1; on \bench-Fast, Gemini-2.5-Flash-Image did so for 14 of 820 prompts and Gemini-3-Pro-Image for 3.
The GPT models always returned an image.
Each such response scores zero, and the result files keep its text.
Of the 62 responses without an image, 48 contain text and 14 are empty.
Three examples follow, with the instructions and responses verbatim.

\begin{abstention}{Gemini-2.5-Flash-Image on \bench-Challenge}
\textbf{Instruction.} ``Draw nine red squares and three blue squares. Use more orange squares than red squares. Arrange all the red squares so they are not all in the same row. Arrange all the red squares in one column. Arrange all the blue squares so they are not all in the same row. Arrange all the blue squares so they are not all in the same column. Make every red square the same size as every blue square. Set the objects against a plain beige background; do not add other colored objects.''
\begin{abstentionresponse}
``I cannot fulfill your request. You have provided contradictory instructions regarding the arrangement of the red squares. You asked that `all the red squares so they are not all in the same row' and also that `all the red squares in one column.' These two statements cannot both be true simultaneously.''
\end{abstentionresponse}
\begin{abstentionwhy}
The two requirements are compatible: nine red squares stacked in one column share a column and, because they are stacked vertically, are not all in the same row.
\end{abstentionwhy}
\end{abstention}

\begin{abstention}{Gemini-3-Pro-Image on \bench-Challenge}
\textbf{Instruction.} ``Show nine orange squares and two pink triangles. Use more cyan squares than orange squares. Arrange all the pink triangles in one row. Arrange all the cyan squares so they are not all in the same row. Arrange all the cyan squares so they are not all in the same column. Make every cyan square the same size as every orange square. Arrange all the cyan squares in row 1, column 1, row 1, column 2, row 1, column 3, row 1, column 4, row 2, column 2, row 2, column 3, row 2, column 4, row 3, column 1, row 3, column 2, and row 3, column 3 of a 3-by-4 grid. Set the objects against a plain pale cyan background; do not add other colored objects.''
\begin{abstentionresponse}
``I cannot generate the image because the constraints are contradictory. The prompt asks for a total of 19 squares (nine orange and ten cyan, as their specific positions are listed) to be arranged within a 3-by-4 grid, which has only 12 available spots.'' (Markdown emphasis removed.)
\end{abstentionresponse}
\begin{abstentionwhy}
Only the ten cyan squares are assigned to grid cells, and ten cells are listed.
The nine orange squares have no grid position and can be placed anywhere outside the grid.
\end{abstentionwhy}
\end{abstention}

\begin{abstention}{Gemini-2.5-Flash-Image on \bench-Challenge}
\textbf{Instruction.} ``Show one pink square, one green circle, nine cyan circles, two yellow circles, and one purple square. Use four times as many cyan squares as yellow circles. Make the pink square contain the green circle. Place all the yellow circles above all the cyan squares. Place the purple square to the left of all the cyan squares. Place all the yellow circles to the right of the purple square. Place all the cyan squares in the right area. Keep the background plain black, with no additional colored objects.''
\begin{abstentionresponse}
``I'm sorry, I cannot fulfill that request. I am unable to keep track of that many rules for generating an image.''
\end{abstentionresponse}
\begin{abstentionwhy}
The task is long but satisfiable: its reference image passes the verifier, and Gemini-3.1-Flash-Image generated an image that satisfies every constraint of the same task.
\end{abstentionwhy}
\end{abstention}

\subsection{Per-constraint pass rates of API models}
\label{sec:api-atomic-details}

We score all 5,040 outputs of the seven API models on \bench-Challenge and record the pass-or-fail decision of every constraint check.
The pass rate of a constraint type pools all of its checks, and responses without an image count as failures.
Table~\ref{tab:vvr-api-constraint-types} gives the pass rate of every constraint type.

\input{tables/vvr_api_constraint_types_challenge}

\subsection{Complexity-matched family analysis}
\label{sec:family-presence-details}

Table~\ref{tab:vvr-family-presence-conditioned} tests whether a model loses accuracy on tasks that contain a constraint family, beyond what the tasks' complexity explains.
For each model and family, it reports \bench accuracy on the tasks that contain the family and, in parentheses, the difference from tasks without the family at matched complexity.
The largest negative differences identify family-specific weaknesses: GPT-Image-1-mini loses 9.7 points on tasks with \SpatialTag{Spatial} constraints, HunyuanImage-2.1 loses 10.7 points with \SizeTag{Size} constraints, and FLUX.2-dev loses 4.0 points with \CardinalityTag{Cardinality} constraints.
Models that solve few tasks show differences near zero.

To match complexity, we stratify \bench prompts by floored integer complexity and, within every stratum that contains tasks with and without family $f$, weight the accuracy of tasks without $f$ by the number of tasks with $f$.
The difference for model $m$ is
\begin{equation}
\Delta_{m,f}=\sum_{c}w_{f,c}\left[
\operatorname{Acc}_{m}(f,c)-\operatorname{Acc}_{m}(\neg f,c)\right],
\end{equation}
where $w_{f,c}$ is the family present complexity distribution.
The matched supports are 3,077 Grounding prompts (96\% coverage), 3,297 Cardinality (98\%), 6,713 Spatial (79\%), 3,182 Size (100\%), and 3,027 Topology (100\%).
Here ``present'' means that the benchmark sampled an explicit constraint from that family; ordinary object realization still appears throughout the benchmark.
The background and forbidden-content constraints apply to every task, so they have no tasks without them to compare against.

\input{tables/vvr_family_presence_conditioned}

Family tags can co-occur.
As a sensitivity check, a linear probability model with all five family indicators and integer complexity fixed effects preserves the largest negative profiles.
GPT-Image-1-mini Spatial changes from $-9.7$ to $-14.6$ points, FLUX.2-dev Cardinality from $-4.0$ to $-5.9$, and HunyuanImage-2.1 Size from $-10.7$ to $-12.6$.
Positive associations for GPT-Image-2 are less stable under this adjustment.

\subsection{Failure examples}
\label{sec:failure-examples}

\begin{figure}[H]
\centering
\begin{minipage}[t]{0.32\linewidth}\centering
\includegraphics[width=\linewidth]{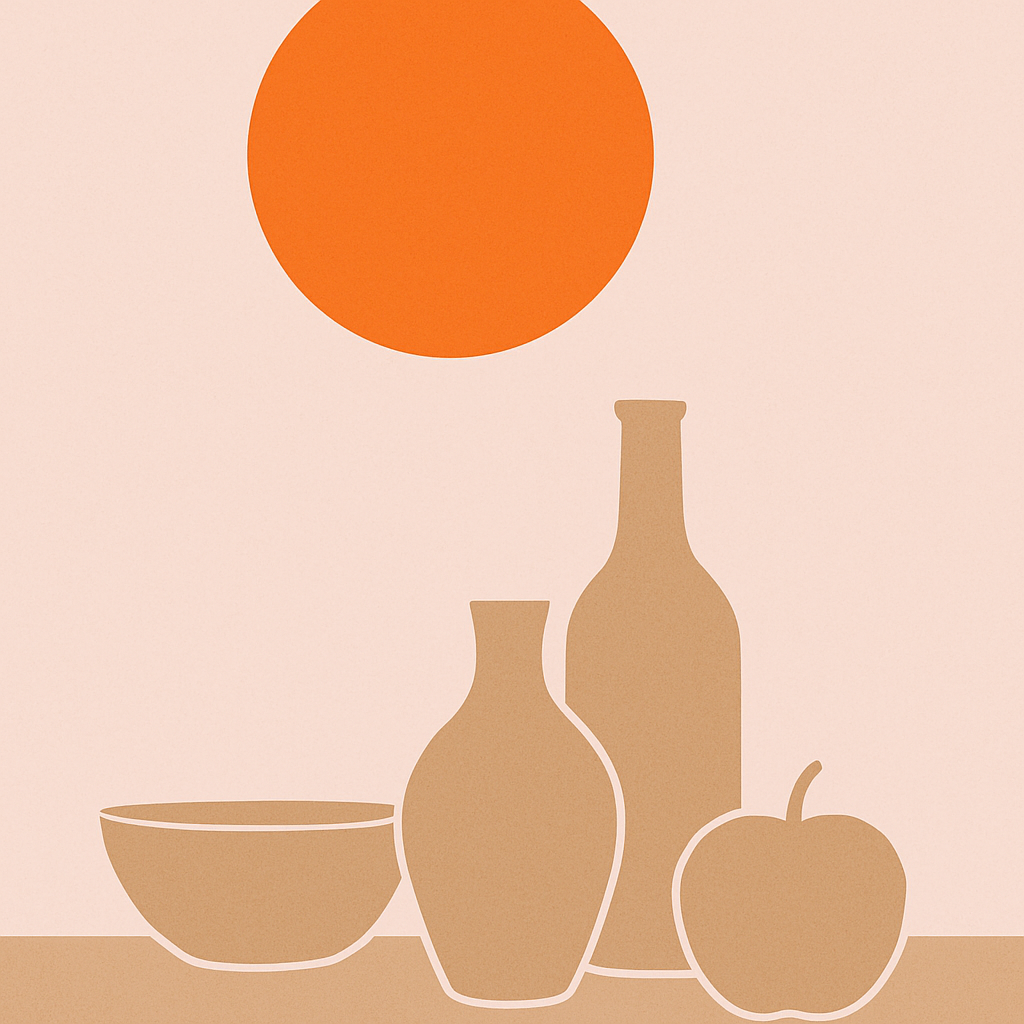}\\
{\scriptsize (a)}
\end{minipage}\hfill
\begin{minipage}[t]{0.32\linewidth}\centering
\includegraphics[width=\linewidth]{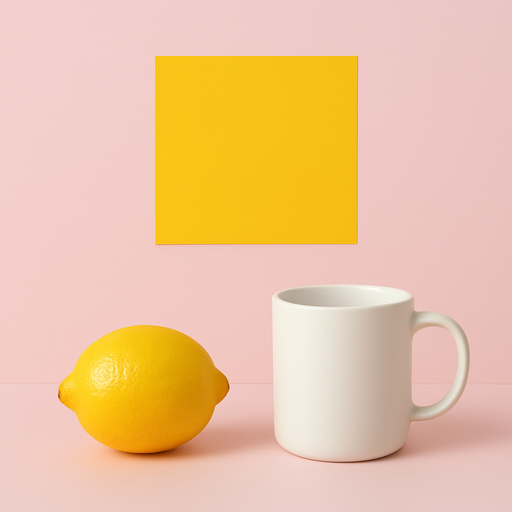}\\
{\scriptsize (b)}
\end{minipage}\hfill
\begin{minipage}[t]{0.32\linewidth}\centering
\includegraphics[width=\linewidth]{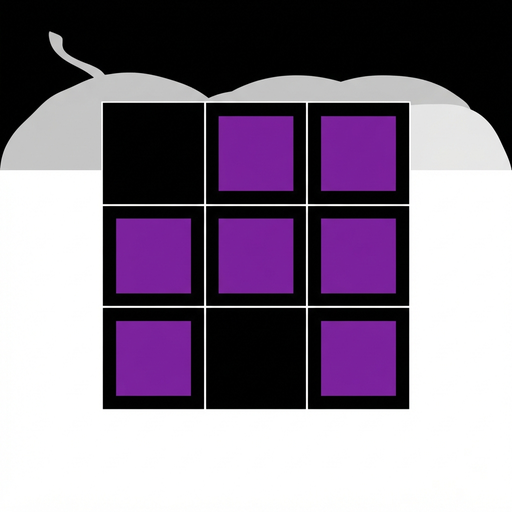}\\
{\scriptsize (c)}
\end{minipage}
\caption{Failed API model outputs on \bench-Fast.
Prompts: (a) ``Place an orange circle in the top area.
Set the objects against a plain pale pink background; do not add other colored objects.'' (b) ``Place a yellow square in the top area.
Set the objects against a plain pale pink background; do not add other colored objects.'' (c) ``Arrange three purple squares in these cells of a 3-by-3 grid: top center, top right, and middle right.
Set the objects against a plain black background; do not add other colored objects.''}
\label{fig:vvr-failure-examples}
\end{figure}

Figure~\ref{fig:vvr-failure-examples} shows three failed \bench-Fast outputs of API models that add content the prompt excludes: vases, a bowl, and an apple; a lemon and a mug; and additional squares and shapes around the grid.

\section{Training Setup}
\label{sec:training-setup}

Table~\ref{tab:sd35-training-reproducibility} reports the settings that define the optimization and reward distribution.
The released resolved configurations and data manifests retain the remaining implementation metadata.

\input{tables/sd35_training_reproducibility}

\subsection{Post-training evaluation}
\label{sec:posttraining-eval-details}

Each model generates one image per prompt with a fixed seed, except on GenEval, which uses four images per prompt.

\paragraph{Training-objective benchmarks.}
Each reward objective is evaluated on its own held-out benchmark.
\bench accuracy uses the 10,000 \bench tasks.
GenEval~\citep{ghosh2023geneval} uses its 553 prompts with four images each, for 2,212 images.
GenEval2~\citep{kamath2025geneval2} uses its fixed 80-prompt held-out split.
OCR~\citep{liu2025flow} uses 1,018 held-out text-rendering prompts scored by normalized edit accuracy.
Each benchmark is reported in its own units.

\paragraph{Preference benchmarks.}
PickScore~\citep{kirstain2023pickscore} is evaluated on the 500 unique prompts of the Pick-a-Pic v1 \texttt{validation\_unique} split.
HPSv2.1~\citep{wu2023human} is evaluated on the complete HPDv2 benchmark, 800 prompts in each of four domains (anime, concept art, paintings, and photo), and reported as the unweighted mean of the four domain means.

\paragraph{Cross-domain panel.}
The remaining metrics use a shared panel of four prompt sets: all 200 DrawBench~\citep{saharia2022imagen} prompts and fixed 1,000-prompt subsets of PartiPrompts~\citep{yu2022parti}, DPG-Bench~\citep{hu2024ella}, and T2I-CompBench~\citep{huang2023t2icompbench}.
On this panel we report HPSv3~\citep{ma2025hpsv3}, which has no canonical prompt benchmark, CLIPScore~\citep{hessel2021clipscore}, LAION aesthetic score~\citep{schuhmann2022aesthetic}, ImageReward~\citep{xu2023imagereward}, and UnifiedReward~\citep{wang2025unifiedreward}.
Each metric is averaged within a prompt set and then across the four sets, so the larger sets do not dominate.
Only the five-reward objective trains on one of these metrics (UnifiedReward); together they test transfer to prompt distributions outside the training tasks.

\section{Complete RLVVR Results}
\label{sec:posttraining-results-full}

\subsection{\texorpdfstring{\bench}{VVRBench} results by complexity}
\label{sec:posttraining-vvr-bench-full}

Table~\ref{tab:vvr-bench-sd35-posttraining} gives the \bench accuracy of every trained model by complexity range; Figure~\ref{fig:vvr-easy-complexity-transfer} plots a subset.

\input{tables/vvr_bench_sd35_posttraining}

\subsection{Partial and joint constraint satisfaction}
\label{sec:partial-joint}

For every \bench task, we compute the mean partial-credit score of its count constraints and of its relations.
A count score is one minus the relative count error, averaged over groups, and a relation score is the mean graded score of the task's relations.
From these we report a partial score, the mean graded score, and the fraction of tasks in which every count or every relation is satisfied (Table~\ref{tab:vvr-partial-joint}).
Relation columns use only the tasks with at least one relation.
Figure~\ref{fig:vvr-partial-joint-gap} expresses VVR-Easy's values as the share of the gap between the pretrained model and VVR-Matched that VVR-Easy closes.
In every range from $C_3$ to $C_5$, VVR-Easy closes more of the gap in partial scores than in the fraction of tasks with every constraint of a kind satisfied, and the difference grows with complexity.

\input{tables/vvr_partial_joint}

\subsection{External task, quality, and alignment metrics}
\label{sec:postfreeze-external-full}

Table~\ref{tab:sd35-postfreeze-external-full} extends Table~\ref{tab:sd35-postfreeze-external} to all nine trained models.

\input{tables/sd35_postfreeze_external_full}

Table~\ref{tab:external-bootstrap} gives paired bootstrap intervals for the mixture comparisons.
The pretrained model's evaluation retained only aggregate scores, so comparisons with it have no intervals.

\input{tables/sd35_external_bootstrap}

\paragraph{Matched-complexity GenEval2 mixture.}
Replacing VVR-Easy with VVR-Matched in the GenEval2 mixture raises \bench accuracy from 21.82\% to 33.50\% and GenEval2 from 0.478 to 0.491; accuracy in $C_3$, $C_4$, and $C_5$ rises from 14.08\%, 6.41\%, and 1.10\% to 32.34\%, 20.32\%, and 9.22\%.
This mixture raises seven of ten non-VVR metrics over GenEval2 alone, with intervals excluding zero for HPSv2.1 ($+0.004$) and ImageReward ($+0.038$).

\section{Human Preference Study}
\label{sec:human-study-details}

Three annotators each compare the same 400 image pairs and choose the image they prefer given the prompt, with a tie option.
The study contains two comparisons, VVR-Easy against the pretrained model and GenEval2 mixed with VVR-Easy against GenEval2, and each of five prompt suites contributes 40 prompts to each comparison.
The 400-prompt study uses 80 unique prompts from each of VVR, GenEval2, GenEval, OCR, and DrawBench.
The VVR prompts were drawn, 16 from each of five complexity bins, from a candidate pool of 11,250 tasks that preceded the final benchmark; 74 of them are \bench tasks, and none appears in VVR-Easy or Challenge.
The GenEval sample is balanced across its six task categories.
Each prompt appears in one comparison, paired generations share a sampling seed, and model identity and left and right order are hidden during annotation.
Each annotator sees the pairs in an independently randomized order and left-right assignment.
Win rates average each prompt's score over the annotators (win 1, tie 0.5, loss 0), and intervals are 95\% bootstrap intervals over prompts.
Each annotator separately favors the VVR-trained model in every suite of both comparisons.
On pairs where both annotators chose an image, the mean pairwise agreement is 83.8\% (84.9\%, 81.9\%, and 84.6\% for the three annotator pairs), and Fleiss' $\kappa$ among the three annotators, with ties as a third label, is 0.49.
Table~\ref{tab:human-annotator-agreement} gives the agreement of each annotator pair.

\input{tables/human_annotator_agreement}

%% file: tables/vvr_bench_verifier_bank.tex
\begin{table}[H]
\centering
\scriptsize
\setlength{\tabcolsep}{3pt}
\renewcommand{\arraystretch}{1.05}
\begin{tabular}{
  >{\raggedright\arraybackslash}p{0.10\textwidth}
  >{\raggedright\arraybackslash}p{0.55\textwidth}
  >{\raggedright\arraybackslash}p{0.25\textwidth}}
\toprule
Family & Exact constraint types \& their supported values & Contribution to $C(s)$ \\
\midrule
Grounding & \texttt{color\_attribute}: red, orange, yellow, green, cyan, blue, purple, pink
& $L(n_i)$ for the referenced group \\
& \texttt{shape\_attribute}: circle, square, triangle
& $L(n_i)$ for the referenced group \\
& \texttt{color\_shape\_binding}
& No additional term; the bound group's color and shape terms already account for it \\
\midrule
Cardinality & \texttt{exact\_count}: 1--10
& $L(n_i)$ for the referenced group \\
& \texttt{same\_count}; \texttt{more\_than\_count}; \texttt{fewer\_than\_count}
& $L(N_r)$ \\
& \texttt{times\_as\_many}: factor $k\in\{2,3,4,5\}$
& $L(N_r)+\log_2 k$, $k\in\{2,3,4,5\}$ \\
\midrule
Spatial & \texttt{absolute\_region}: top left, top, top right, left, center, right, bottom left, bottom, bottom right
& $L(N_l)$ \\
& \texttt{grid\_occupancy}: cells of a $2\times2$, $2\times3$, or $3\times3$ grid ($3\times4$ in \bench-Challenge)
& $L(N_l)$ \\
& \texttt{left\_of}; \texttt{right\_of}; \texttt{above}; \texttt{below};
  \texttt{all\_left\_of}$^\dagger$; \texttt{all\_right\_of}$^\dagger$; \texttt{all\_above}$^\dagger$;
  \texttt{all\_below}$^\dagger$; \texttt{leftmost}; \texttt{rightmost};
  \texttt{topmost}; \texttt{bottommost}; \texttt{between};
  \texttt{same\_row}; \texttt{same\_column}; \texttt{all\_same\_row}$^\dagger$;
  \texttt{all\_same\_column}$^\dagger$; \texttt{not\_all\_same\_row}$^\dagger$;
  \texttt{not\_all\_same\_column}$^\dagger$; \texttt{closer\_than};
  \texttt{farther\_than}
& $L(N_r)$ \\
\midrule
Size & \texttt{larger\_than}; \texttt{smaller\_than}; \texttt{same\_size};
  \texttt{all\_larger\_than}$^\dagger$; \texttt{all\_smaller\_than}$^\dagger$;
  \texttt{all\_same\_size}$^\dagger$; \texttt{largest}; \texttt{smallest}
& $L(N_r)$ \\
& \texttt{not\_all\_same\_size}$^\dagger$
& $2\log_2 N_r$ \\
\midrule
Topology & \texttt{touching}; \texttt{not\_touching}; \texttt{inside};
  \texttt{contains}
& $L(N_r)$ \\
& \texttt{each\_inside}$^\dagger$; \texttt{each\_contains}$^\dagger$
& $m_r$ \\
\bottomrule
\end{tabular}
\caption{The exact 46 constraint types, their supported values, and their contributions to structural complexity.
$^\dagger$ marks types that appear only in \bench-Challenge.
Types with the same cost rule share a row, and types without listed values take only object groups as arguments.
The background constraint supports white, black, light gray, dark gray, beige, pale pink, and pale cyan.
The background and forbidden-content constraints in $\mathcal{B}$ and $\mathcal{F}$ apply to every task and do not contribute to $C(s)$.}
\label{tab:vvr-verifier-bank}
\end{table}

%% file: tables/vvr_bench_core_api_820.tex
\begin{table}[H]
\centering
\scriptsize
\setlength{\tabcolsep}{2.5pt}
\caption{Accuracy (\%) of API models on \bench-Fast, an 820 task subset of \bench with 20 tasks at each attainable integer complexity from 3 to 44.
The GPT-Image-2 models exceed 80\% overall but fall to 51\% to 57\% in $C_5$.
Gemini models reach 37\% to 47\%.
Subscripts are 95\% confidence margins.}
\label{tab:vvr-bench-core-api-820}
\begin{tabular}{lrrrrrr}
\toprule
Model & Accuracy (\%) $\uparrow$ & 3 to 10 & 11 to 18 & 19 to 26 & 27 to 35 & 36 to 44 \\
\midrule
GPT-Image-2.5-Sunburst & \textbf{84.51}$_{\pm 2.64}$ & \textbf{100.00}$_{\pm 2.67}$ & \textbf{98.75}$_{\pm 3.19}$ & \textbf{96.25}$_{\pm 4.19}$ & \textbf{77.22}$_{\pm 6.66}$ & \textbf{56.67}$_{\pm 7.30}$ \\
GPT-Image-2 & 82.20$_{\pm 2.77}$ & 98.57$_{\pm 3.63}$ & \textbf{98.75}$_{\pm 3.19}$ & 95.62$_{\pm 4.38}$ & 74.44$_{\pm 6.84}$ & 50.56$_{\pm 7.24}$ \\
Gemini-3.1-Flash-Image & 47.20$_{\pm 3.42}$ & 85.71$_{\pm 6.75}$ & 60.00$_{\pm 7.74}$ & 51.25$_{\pm 7.68}$ & 30.56$_{\pm 7.08}$ & 18.89$_{\pm 6.35}$ \\
Gemini-2.5-Flash-Image & 45.61$_{\pm 3.42}$ & 76.43$_{\pm 7.68}$ & 65.62$_{\pm 7.65}$ & 51.25$_{\pm 7.68}$ & 31.11$_{\pm 7.10}$ & 13.33$_{\pm 5.74}$ \\
Gemini-3.1-Flash-Lite-Image & 42.07$_{\pm 3.41}$ & 85.71$_{\pm 6.75}$ & 56.25$_{\pm 7.74}$ & 40.00$_{\pm 7.74}$ & 25.00$_{\pm 6.80}$ & 14.44$_{\pm 5.88}$ \\
Gemini-3-Pro-Image & 37.20$_{\pm 3.36}$ & 62.14$_{\pm 8.26}$ & 52.50$_{\pm 7.71}$ & 38.75$_{\pm 7.73}$ & 26.67$_{\pm 6.90}$ & 13.33$_{\pm 5.74}$ \\
GPT-Image-1-mini & 32.44$_{\pm 3.28}$ & 92.86$_{\pm 5.51}$ & 60.00$_{\pm 7.74}$ & 16.88$_{\pm 6.56}$ & 4.44$_{\pm 4.08}$ & 2.78$_{\pm 3.56}$ \\
\bottomrule
\end{tabular}
\end{table}

%% file: tables/vvr_api_constraint_types_challenge.tex
\begin{table}[t]
\centering
\scriptsize
\setlength{\tabcolsep}{3pt}
\renewcommand{\arraystretch}{0.92}
\caption{Pass rates (\%) of API models for every constraint type on \bench-Challenge.
$n$ is the number of checks of each type per model, and Avg is the unweighted mean over the seven models.
Responses without an image count as failures.
Within each family, types are sorted by Avg.
Cell shading is proportional to the pass rate.}
\label{tab:vvr-api-constraint-types}
\resizebox{\linewidth}{!}{%
\begin{tabular}{@{}lrcccccccc@{}}
\toprule
Constraint type & $n$ & \shortstack{GPT-Image-\\2.5-Sunburst} & \shortstack{GPT-\\Image-2} & \shortstack{Gemini-3.1-\\Flash-Lite} & \shortstack{Gemini-\\3-Pro} & \shortstack{Gemini-\\3.1-Flash} & \shortstack{Gemini-\\2.5-Flash} & \shortstack{GPT-Image-\\1-mini} & Avg \\
\midrule
\multicolumn{10}{@{}l}{\GroundingTag{Grounding}} \\
\quad\texttt{color\_shape\_binding} & 3794 & \cellcolor{vvrHigh!54!white}99 & \cellcolor{vvrHigh!54!white}98 & \cellcolor{vvrHigh!50!white}91 & \cellcolor{vvrHigh!49!white}89 & \cellcolor{vvrHigh!50!white}92 & \cellcolor{vvrHigh!46!white}84 & \cellcolor{vvrHigh!50!white}92 & \cellcolor{vvrHigh!51!white}\textbf{92} \\
\quad\texttt{shape\_attribute} & 3794 & \cellcolor{vvrHigh!54!white}99 & \cellcolor{vvrHigh!54!white}98 & \cellcolor{vvrHigh!52!white}94 & \cellcolor{vvrHigh!52!white}94 & \cellcolor{vvrHigh!52!white}94 & \cellcolor{vvrHigh!48!white}88 & \cellcolor{vvrHigh!51!white}94 & \cellcolor{vvrHigh!52!white}\textbf{94} \\
\quad\texttt{color\_attribute} & 3794 & \cellcolor{vvrHigh!55!white}99 & \cellcolor{vvrHigh!54!white}99 & \cellcolor{vvrHigh!54!white}98 & \cellcolor{vvrHigh!55!white}99 & \cellcolor{vvrHigh!54!white}98 & \cellcolor{vvrHigh!52!white}94 & \cellcolor{vvrHigh!52!white}95 & \cellcolor{vvrHigh!54!white}\textbf{98} \\
\midrule
\multicolumn{10}{@{}l}{\CardinalityTag{Cardinality}} \\
\quad\texttt{same\_count} & 266 & \cellcolor{vvrHigh!31!white}56 & \cellcolor{vvrHigh!24!white}43 & \cellcolor{vvrHigh!17!white}32 & \cellcolor{vvrHigh!15!white}27 & \cellcolor{vvrHigh!15!white}28 & \cellcolor{vvrHigh!9!white}17 & \cellcolor{vvrHigh!9!white}16 & \cellcolor{vvrHigh!17!white}\textbf{31} \\
\quad\texttt{times\_as\_many} & 136 & \cellcolor{vvrHigh!32!white}58 & \cellcolor{vvrHigh!24!white}44 & \cellcolor{vvrHigh!21!white}39 & \cellcolor{vvrHigh!19!white}35 & \cellcolor{vvrHigh!19!white}34 & \cellcolor{vvrHigh!9!white}16 & \cellcolor{vvrHigh!4!white}7 & \cellcolor{vvrHigh!18!white}\textbf{33} \\
\quad\texttt{fewer\_than\_count} & 20 & \cellcolor{vvrHigh!30!white}55 & \cellcolor{vvrHigh!30!white}55 & \cellcolor{vvrHigh!25!white}45 & \cellcolor{vvrHigh!30!white}55 & \cellcolor{vvrHigh!22!white}40 & \cellcolor{vvrHigh!36!white}65 & \cellcolor{vvrHigh!16!white}30 & \cellcolor{vvrHigh!27!white}\textbf{49} \\
\quad\texttt{more\_than\_count} & 20 & \cellcolor{vvrHigh!33!white}60 & \cellcolor{vvrHigh!38!white}70 & \cellcolor{vvrHigh!28!white}50 & \cellcolor{vvrHigh!36!white}65 & \cellcolor{vvrHigh!38!white}70 & \cellcolor{vvrHigh!22!white}40 & \cellcolor{vvrHigh!22!white}40 & \cellcolor{vvrHigh!31!white}\textbf{56} \\
\quad\texttt{exact\_count} & 3794 & \cellcolor{vvrHigh!45!white}81 & \cellcolor{vvrHigh!41!white}75 & \cellcolor{vvrHigh!39!white}72 & \cellcolor{vvrHigh!35!white}64 & \cellcolor{vvrHigh!34!white}62 & \cellcolor{vvrHigh!27!white}49 & \cellcolor{vvrHigh!27!white}49 & \cellcolor{vvrHigh!35!white}\textbf{64} \\
\midrule
\multicolumn{10}{@{}l}{\SpatialTag{Spatial}} \\
\quad\texttt{grid\_occupancy} & 20 & \cellcolor{vvrHigh!6!white}10 & \cellcolor{vvrHigh!6!white}10 & \cellcolor{vvrHigh!11!white}20 & \cellcolor{vvrHigh!6!white}10 & \cellcolor{vvrHigh!0!white}0 & \cellcolor{vvrHigh!0!white}0 & \cellcolor{vvrHigh!0!white}0 & \cellcolor{vvrHigh!4!white}\textbf{7} \\
\quad\texttt{rightmost} & 20 & \cellcolor{vvrHigh!22!white}40 & \cellcolor{vvrHigh!16!white}30 & \cellcolor{vvrHigh!30!white}55 & \cellcolor{vvrHigh!28!white}50 & \cellcolor{vvrHigh!33!white}60 & \cellcolor{vvrHigh!11!white}20 & \cellcolor{vvrHigh!16!white}30 & \cellcolor{vvrHigh!22!white}\textbf{41} \\
\quad\texttt{leftmost} & 20 & \cellcolor{vvrHigh!8!white}15 & \cellcolor{vvrHigh!30!white}55 & \cellcolor{vvrHigh!30!white}55 & \cellcolor{vvrHigh!41!white}75 & \cellcolor{vvrHigh!30!white}55 & \cellcolor{vvrHigh!16!white}30 & \cellcolor{vvrHigh!22!white}40 & \cellcolor{vvrHigh!26!white}\textbf{46} \\
\quad\texttt{bottommost} & 20 & \cellcolor{vvrHigh!41!white}75 & \cellcolor{vvrHigh!41!white}75 & \cellcolor{vvrHigh!30!white}55 & \cellcolor{vvrHigh!22!white}40 & \cellcolor{vvrHigh!30!white}55 & \cellcolor{vvrHigh!22!white}40 & \cellcolor{vvrHigh!22!white}40 & \cellcolor{vvrHigh!30!white}\textbf{54} \\
\quad\texttt{between} & 20 & \cellcolor{vvrHigh!47!white}85 & \cellcolor{vvrHigh!33!white}60 & \cellcolor{vvrHigh!25!white}45 & \cellcolor{vvrHigh!30!white}55 & \cellcolor{vvrHigh!33!white}60 & \cellcolor{vvrHigh!28!white}50 & \cellcolor{vvrHigh!16!white}30 & \cellcolor{vvrHigh!30!white}\textbf{55} \\
\quad\texttt{topmost} & 20 & \cellcolor{vvrHigh!44!white}80 & \cellcolor{vvrHigh!47!white}85 & \cellcolor{vvrHigh!30!white}55 & \cellcolor{vvrHigh!36!white}65 & \cellcolor{vvrHigh!28!white}50 & \cellcolor{vvrHigh!25!white}45 & \cellcolor{vvrHigh!25!white}45 & \cellcolor{vvrHigh!33!white}\textbf{61} \\
\quad\texttt{all\_same\_column} & 20 & \cellcolor{vvrHigh!55!white}100 & \cellcolor{vvrHigh!52!white}95 & \cellcolor{vvrHigh!44!white}80 & \cellcolor{vvrHigh!28!white}50 & \cellcolor{vvrHigh!33!white}60 & \cellcolor{vvrHigh!28!white}50 & \cellcolor{vvrHigh!33!white}60 & \cellcolor{vvrHigh!39!white}\textbf{71} \\
\quad\texttt{all\_right\_of} & 122 & \cellcolor{vvrHigh!47!white}86 & \cellcolor{vvrHigh!43!white}78 & \cellcolor{vvrHigh!43!white}78 & \cellcolor{vvrHigh!45!white}82 & \cellcolor{vvrHigh!41!white}74 & \cellcolor{vvrHigh!38!white}69 & \cellcolor{vvrHigh!30!white}55 & \cellcolor{vvrHigh!41!white}\textbf{74} \\
\quad\texttt{all\_left\_of} & 281 & \cellcolor{vvrHigh!49!white}88 & \cellcolor{vvrHigh!43!white}79 & \cellcolor{vvrHigh!43!white}78 & \cellcolor{vvrHigh!46!white}83 & \cellcolor{vvrHigh!41!white}75 & \cellcolor{vvrHigh!39!white}72 & \cellcolor{vvrHigh!31!white}57 & \cellcolor{vvrHigh!42!white}\textbf{76} \\
\quad\texttt{closer\_than} & 40 & \cellcolor{vvrHigh!51!white}92 & \cellcolor{vvrHigh!54!white}98 & \cellcolor{vvrHigh!47!white}85 & \cellcolor{vvrHigh!48!white}88 & \cellcolor{vvrHigh!38!white}70 & \cellcolor{vvrHigh!37!white}68 & \cellcolor{vvrHigh!32!white}57 & \cellcolor{vvrHigh!44!white}\textbf{80} \\
\quad\texttt{farther\_than} & 20 & \cellcolor{vvrHigh!52!white}95 & \cellcolor{vvrHigh!50!white}90 & \cellcolor{vvrHigh!47!white}85 & \cellcolor{vvrHigh!44!white}80 & \cellcolor{vvrHigh!44!white}80 & \cellcolor{vvrHigh!33!white}60 & \cellcolor{vvrHigh!44!white}80 & \cellcolor{vvrHigh!45!white}\textbf{81} \\
\quad\texttt{right\_of} & 20 & \cellcolor{vvrHigh!55!white}100 & \cellcolor{vvrHigh!52!white}95 & \cellcolor{vvrHigh!50!white}90 & \cellcolor{vvrHigh!47!white}85 & \cellcolor{vvrHigh!47!white}85 & \cellcolor{vvrHigh!38!white}70 & \cellcolor{vvrHigh!38!white}70 & \cellcolor{vvrHigh!47!white}\textbf{85} \\
\quad\texttt{all\_same\_row} & 32 & \cellcolor{vvrHigh!53!white}97 & \cellcolor{vvrHigh!52!white}94 & \cellcolor{vvrHigh!46!white}84 & \cellcolor{vvrHigh!43!white}78 & \cellcolor{vvrHigh!52!white}94 & \cellcolor{vvrHigh!48!white}88 & \cellcolor{vvrHigh!43!white}78 & \cellcolor{vvrHigh!48!white}\textbf{88} \\
\quad\texttt{absolute\_region} & 384 & \cellcolor{vvrHigh!54!white}98 & \cellcolor{vvrHigh!53!white}97 & \cellcolor{vvrHigh!50!white}91 & \cellcolor{vvrHigh!51!white}92 & \cellcolor{vvrHigh!50!white}90 & \cellcolor{vvrHigh!42!white}77 & \cellcolor{vvrHigh!44!white}80 & \cellcolor{vvrHigh!49!white}\textbf{89} \\
\quad\texttt{all\_below} & 57 & \cellcolor{vvrHigh!54!white}98 & \cellcolor{vvrHigh!53!white}96 & \cellcolor{vvrHigh!49!white}89 & \cellcolor{vvrHigh!50!white}91 & \cellcolor{vvrHigh!48!white}88 & \cellcolor{vvrHigh!45!white}82 & \cellcolor{vvrHigh!44!white}81 & \cellcolor{vvrHigh!49!white}\textbf{89} \\
\quad\texttt{all\_above} & 102 & \cellcolor{vvrHigh!55!white}100 & \cellcolor{vvrHigh!54!white}99 & \cellcolor{vvrHigh!50!white}91 & \cellcolor{vvrHigh!51!white}93 & \cellcolor{vvrHigh!50!white}91 & \cellcolor{vvrHigh!45!white}82 & \cellcolor{vvrHigh!45!white}82 & \cellcolor{vvrHigh!50!white}\textbf{91} \\
\quad\texttt{same\_column} & 20 & \cellcolor{vvrHigh!55!white}100 & \cellcolor{vvrHigh!52!white}95 & \cellcolor{vvrHigh!52!white}95 & \cellcolor{vvrHigh!52!white}95 & \cellcolor{vvrHigh!55!white}100 & \cellcolor{vvrHigh!50!white}90 & \cellcolor{vvrHigh!36!white}65 & \cellcolor{vvrHigh!50!white}\textbf{91} \\
\quad\texttt{not\_all\_same\_row} & 136 & \cellcolor{vvrHigh!54!white}98 & \cellcolor{vvrHigh!53!white}96 & \cellcolor{vvrHigh!53!white}96 & \cellcolor{vvrHigh!50!white}90 & \cellcolor{vvrHigh!52!white}94 & \cellcolor{vvrHigh!47!white}85 & \cellcolor{vvrHigh!50!white}91 & \cellcolor{vvrHigh!51!white}\textbf{93} \\
\quad\texttt{same\_row} & 20 & \cellcolor{vvrHigh!55!white}100 & \cellcolor{vvrHigh!55!white}100 & \cellcolor{vvrHigh!55!white}100 & \cellcolor{vvrHigh!52!white}95 & \cellcolor{vvrHigh!50!white}90 & \cellcolor{vvrHigh!50!white}90 & \cellcolor{vvrHigh!44!white}80 & \cellcolor{vvrHigh!51!white}\textbf{94} \\
\quad\texttt{not\_all\_same\_column} & 145 & \cellcolor{vvrHigh!55!white}99 & \cellcolor{vvrHigh!54!white}98 & \cellcolor{vvrHigh!53!white}97 & \cellcolor{vvrHigh!50!white}91 & \cellcolor{vvrHigh!52!white}95 & \cellcolor{vvrHigh!47!white}85 & \cellcolor{vvrHigh!49!white}90 & \cellcolor{vvrHigh!51!white}\textbf{94} \\
\quad\texttt{below} & 20 & \cellcolor{vvrHigh!52!white}95 & \cellcolor{vvrHigh!52!white}95 & \cellcolor{vvrHigh!55!white}100 & \cellcolor{vvrHigh!52!white}95 & \cellcolor{vvrHigh!55!white}100 & \cellcolor{vvrHigh!41!white}75 & \cellcolor{vvrHigh!55!white}100 & \cellcolor{vvrHigh!52!white}\textbf{94} \\
\quad\texttt{left\_of} & 20 & \cellcolor{vvrHigh!55!white}100 & \cellcolor{vvrHigh!55!white}100 & \cellcolor{vvrHigh!52!white}95 & \cellcolor{vvrHigh!52!white}95 & \cellcolor{vvrHigh!47!white}85 & \cellcolor{vvrHigh!52!white}95 & \cellcolor{vvrHigh!52!white}95 & \cellcolor{vvrHigh!52!white}\textbf{95} \\
\quad\texttt{above} & 20 & \cellcolor{vvrHigh!55!white}100 & \cellcolor{vvrHigh!55!white}100 & \cellcolor{vvrHigh!52!white}95 & \cellcolor{vvrHigh!55!white}100 & \cellcolor{vvrHigh!55!white}100 & \cellcolor{vvrHigh!41!white}75 & \cellcolor{vvrHigh!52!white}95 & \cellcolor{vvrHigh!52!white}\textbf{95} \\
\midrule
\multicolumn{10}{@{}l}{\SizeTag{Size}} \\
\quad\texttt{smallest} & 20 & \cellcolor{vvrHigh!41!white}75 & \cellcolor{vvrHigh!44!white}80 & \cellcolor{vvrHigh!3!white}5 & \cellcolor{vvrHigh!6!white}10 & \cellcolor{vvrHigh!8!white}15 & \cellcolor{vvrHigh!11!white}20 & \cellcolor{vvrHigh!3!white}5 & \cellcolor{vvrHigh!16!white}\textbf{30} \\
\quad\texttt{all\_same\_size} & 487 & \cellcolor{vvrHigh!41!white}74 & \cellcolor{vvrHigh!27!white}50 & \cellcolor{vvrHigh!27!white}50 & \cellcolor{vvrHigh!24!white}44 & \cellcolor{vvrHigh!27!white}48 & \cellcolor{vvrHigh!17!white}31 & \cellcolor{vvrHigh!17!white}32 & \cellcolor{vvrHigh!26!white}\textbf{47} \\
\quad\texttt{same\_size} & 88 & \cellcolor{vvrHigh!54!white}98 & \cellcolor{vvrHigh!50!white}91 & \cellcolor{vvrHigh!49!white}89 & \cellcolor{vvrHigh!47!white}85 & \cellcolor{vvrHigh!44!white}80 & \cellcolor{vvrHigh!38!white}68 & \cellcolor{vvrHigh!39!white}70 & \cellcolor{vvrHigh!46!white}\textbf{83} \\
\quad\texttt{all\_larger\_than} & 264 & \cellcolor{vvrHigh!55!white}99 & \cellcolor{vvrHigh!53!white}96 & \cellcolor{vvrHigh!48!white}88 & \cellcolor{vvrHigh!46!white}83 & \cellcolor{vvrHigh!44!white}80 & \cellcolor{vvrHigh!39!white}70 & \cellcolor{vvrHigh!50!white}92 & \cellcolor{vvrHigh!48!white}\textbf{87} \\
\quad\texttt{all\_smaller\_than} & 20 & \cellcolor{vvrHigh!55!white}100 & \cellcolor{vvrHigh!55!white}100 & \cellcolor{vvrHigh!52!white}95 & \cellcolor{vvrHigh!41!white}75 & \cellcolor{vvrHigh!47!white}85 & \cellcolor{vvrHigh!44!white}80 & \cellcolor{vvrHigh!44!white}80 & \cellcolor{vvrHigh!48!white}\textbf{88} \\
\quad\texttt{smaller\_than} & 20 & \cellcolor{vvrHigh!55!white}100 & \cellcolor{vvrHigh!55!white}100 & \cellcolor{vvrHigh!47!white}85 & \cellcolor{vvrHigh!50!white}90 & \cellcolor{vvrHigh!52!white}95 & \cellcolor{vvrHigh!41!white}75 & \cellcolor{vvrHigh!50!white}90 & \cellcolor{vvrHigh!50!white}\textbf{91} \\
\quad\texttt{largest} & 20 & \cellcolor{vvrHigh!55!white}100 & \cellcolor{vvrHigh!55!white}100 & \cellcolor{vvrHigh!47!white}85 & \cellcolor{vvrHigh!55!white}100 & \cellcolor{vvrHigh!50!white}90 & \cellcolor{vvrHigh!41!white}75 & \cellcolor{vvrHigh!50!white}90 & \cellcolor{vvrHigh!50!white}\textbf{91} \\
\quad\texttt{larger\_than} & 33 & \cellcolor{vvrHigh!53!white}97 & \cellcolor{vvrHigh!53!white}97 & \cellcolor{vvrHigh!52!white}94 & \cellcolor{vvrHigh!52!white}94 & \cellcolor{vvrHigh!52!white}94 & \cellcolor{vvrHigh!38!white}70 & \cellcolor{vvrHigh!53!white}97 & \cellcolor{vvrHigh!50!white}\textbf{92} \\
\quad\texttt{not\_all\_same\_size} & 20 & \cellcolor{vvrHigh!55!white}100 & \cellcolor{vvrHigh!55!white}100 & \cellcolor{vvrHigh!52!white}95 & \cellcolor{vvrHigh!55!white}100 & \cellcolor{vvrHigh!50!white}90 & \cellcolor{vvrHigh!44!white}80 & \cellcolor{vvrHigh!44!white}80 & \cellcolor{vvrHigh!51!white}\textbf{92} \\
\midrule
\multicolumn{10}{@{}l}{\TopologyTag{Topology}} \\
\quad\texttt{each\_contains} & 581 & \cellcolor{vvrHigh!15!white}27 & \cellcolor{vvrHigh!10!white}19 & \cellcolor{vvrHigh!33!white}61 & \cellcolor{vvrHigh!29!white}53 & \cellcolor{vvrHigh!24!white}44 & \cellcolor{vvrHigh!10!white}18 & \cellcolor{vvrHigh!9!white}17 & \cellcolor{vvrHigh!19!white}\textbf{34} \\
\quad\texttt{touching} & 20 & \cellcolor{vvrHigh!28!white}50 & \cellcolor{vvrHigh!38!white}70 & \cellcolor{vvrHigh!33!white}60 & \cellcolor{vvrHigh!28!white}50 & \cellcolor{vvrHigh!38!white}70 & \cellcolor{vvrHigh!11!white}20 & \cellcolor{vvrHigh!0!white}0 & \cellcolor{vvrHigh!25!white}\textbf{46} \\
\quad\texttt{each\_inside} & 138 & \cellcolor{vvrHigh!51!white}92 & \cellcolor{vvrHigh!46!white}83 & \cellcolor{vvrHigh!34!white}62 & \cellcolor{vvrHigh!33!white}60 & \cellcolor{vvrHigh!29!white}54 & \cellcolor{vvrHigh!21!white}38 & \cellcolor{vvrHigh!26!white}47 & \cellcolor{vvrHigh!34!white}\textbf{62} \\
\quad\texttt{contains} & 20 & \cellcolor{vvrHigh!55!white}100 & \cellcolor{vvrHigh!55!white}100 & \cellcolor{vvrHigh!52!white}95 & \cellcolor{vvrHigh!52!white}95 & \cellcolor{vvrHigh!50!white}90 & \cellcolor{vvrHigh!33!white}60 & \cellcolor{vvrHigh!55!white}100 & \cellcolor{vvrHigh!50!white}\textbf{91} \\
\quad\texttt{inside} & 20 & \cellcolor{vvrHigh!52!white}95 & \cellcolor{vvrHigh!55!white}100 & \cellcolor{vvrHigh!55!white}100 & \cellcolor{vvrHigh!55!white}100 & \cellcolor{vvrHigh!47!white}85 & \cellcolor{vvrHigh!38!white}70 & \cellcolor{vvrHigh!52!white}95 & \cellcolor{vvrHigh!51!white}\textbf{92} \\
\quad\texttt{not\_touching} & 259 & \cellcolor{vvrHigh!55!white}100 & \cellcolor{vvrHigh!55!white}100 & \cellcolor{vvrHigh!54!white}99 & \cellcolor{vvrHigh!55!white}100 & \cellcolor{vvrHigh!55!white}99 & \cellcolor{vvrHigh!49!white}89 & \cellcolor{vvrHigh!53!white}97 & \cellcolor{vvrHigh!54!white}\textbf{98} \\
\midrule
\multicolumn{10}{@{}l}{\SceneCheckTag{Background and forbidden content}} \\
\quad\texttt{no\_unrequested\_objects} & 720 & \cellcolor{vvrHigh!34!white}62 & \cellcolor{vvrHigh!28!white}51 & \cellcolor{vvrHigh!31!white}56 & \cellcolor{vvrHigh!22!white}40 & \cellcolor{vvrHigh!19!white}35 & \cellcolor{vvrHigh!16!white}30 & \cellcolor{vvrHigh!26!white}46 & \cellcolor{vvrHigh!25!white}\textbf{46} \\
\quad\texttt{background\_color} & 720 & \cellcolor{vvrHigh!55!white}100 & \cellcolor{vvrHigh!55!white}100 & \cellcolor{vvrHigh!53!white}96 & \cellcolor{vvrHigh!52!white}94 & \cellcolor{vvrHigh!51!white}93 & \cellcolor{vvrHigh!51!white}93 & \cellcolor{vvrHigh!54!white}99 & \cellcolor{vvrHigh!53!white}\textbf{96} \\
\bottomrule
\end{tabular}%
}
\end{table}

%% file: tables/vvr_family_presence_conditioned.tex
\begin{table}[t]
\centering
\scriptsize
\setlength{\tabcolsep}{3.2pt}
\caption{\bench accuracy (\%) on prompts that contain each constraint family.
In parentheses is the difference from prompts without that family at matched complexity.
Models show distinct weaknesses: Spatial for GPT-Image-1-mini ($-9.7$), Size for HunyuanImage 2.1 ($-10.7$), and Cardinality for FLUX.2 dev ($-4.0$).}
\label{tab:vvr-family-presence-conditioned}
\begin{tabular}{lrrrrrr}
\toprule
Model & Overall & Grounding & Cardinality & Spatial & Size & Topology \\
\midrule
GPT-Image-2 & 86.9 & 86.8 $(+0.5)$ & 80.0 $(+8.0)$ & 85.4 $(-4.8)$ & 87.6 $(+10.0)$ & 85.6 $(+7.7)$ \\
GPT-Image-1-mini & 26.4 & 26.6 $(+0.5)$ & 14.2 $(+1.8)$ & 20.7 $(-9.7)$ & 19.2 $(+2.5)$ & 10.7 $(-3.3)$ \\
FLUX.2-dev & 19.1 & 19.0 $(-0.5)$ & 8.5 $(-4.0)$ & 16.3 $(+0.3)$ & 11.7 $(-1.4)$ & 9.1 $(-1.2)$ \\
HunyuanImage-2.1 & 18.8 & 19.0 $(+0.6)$ & 10.3 $(+0.0)$ & 16.5 $(-0.4)$ & 6.5 $(-10.7)$ & 11.5 $(-0.7)$ \\
Qwen-Image-2512 & 5.8 & 6.7 $(+1.0)$ & 2.2 $(-0.7)$ & 4.4 $(+0.2)$ & 2.4 $(-1.3)$ & 3.3 $(+1.1)$ \\
HiDream-I1-Full & 4.2 & 5.3 $(+0.1)$ & 2.0 $(+0.1)$ & 2.6 $(-1.8)$ & 1.6 $(-0.5)$ & 1.5 $(+0.0)$ \\
FLUX.1-dev & 3.9 & 5.2 $(+0.4)$ & 1.8 $(+0.2)$ & 2.6 $(-0.0)$ & 1.1 $(-0.9)$ & 1.8 $(+0.6)$ \\
FLUX.1-schnell & 2.9 & 3.8 $(-0.1)$ & 1.4 $(+0.3)$ & 1.8 $(-0.1)$ & 0.4 $(-1.0)$ & 1.0 $(+0.3)$ \\
SD3.5 Medium & 2.8 & 4.1 $(-0.0)$ & 1.3 $(+0.0)$ & 1.5 $(-0.9)$ & 0.9 $(-0.1)$ & 0.9 $(+0.1)$ \\
SD3.5 Large & 2.5 & 3.5 $(+0.3)$ & 1.7 $(+0.4)$ & 1.3 $(-0.7)$ & 0.5 $(-0.5)$ & 1.0 $(+0.4)$ \\
SDXL 1.0 & 0.0 & 0.1 $(+0.0)$ & 0.0 $(+0.0)$ & 0.0 $(+0.0)$ & 0.0 $(+0.0)$ & 0.0 $(+0.0)$ \\
Sana 1.6B & 0.0 & 0.0 $(+0.0)$ & 0.0 $(+0.0)$ & 0.0 $(+0.0)$ & 0.0 $(+0.0)$ & 0.0 $(+0.0)$ \\
\bottomrule
\end{tabular}
\end{table}

%% file: tables/sd35_training_reproducibility.tex
\begin{table}[H]
\centering
\small
\setlength{\tabcolsep}{5pt}
\renewcommand{\arraystretch}{1.08}
\caption{Reproducible configuration for the final SD3.5 Medium post-training experiments.
Reward proportions are fractions of prompt groups in each update; every rollout is scored only by the reward attached to its prompt.}
\label{tab:sd35-training-reproducibility}
\begin{tabular}{p{0.23\textwidth}p{0.70\textwidth}}
\toprule
Setting & Value \\
\midrule
Trainable parameters
& LoRA on the eight attention projections
\texttt{add\_k}, \texttt{add\_q}, \texttt{add\_v}, \texttt{add\_out},
\texttt{k}, \texttt{q}, \texttt{v}, and \texttt{out}; rank 32 and $\alpha=64$. \\
Generation
& $512\!\times\!512$ pixels; 25 denoising steps; classifier-free guidance
4.5; Gaussian sampling noise with level 0.7. \\
Rollout batch
& 32 prompt groups per update, 24 rollouts per prompt, and 768 generated
images per update. \\
Flow GRPO
& One inner epoch; advantages centered within each prompt group and divided by
the standard deviation over the complete rollout batch; advantages clipped to
$[-5,5]$; policy-ratio clip $10^{-4}$; KL coefficient 0.04. The first 24 of
the 25 sampled transitions contribute to the update. \\
Optimization
& AdamW; learning rate $3\!\times\!10^{-4}$; $\beta_1=0.9$,
$\beta_2=0.999$; $\epsilon=10^{-8}$; weight decay $10^{-4}$; maximum gradient
norm 1.0; FP16 mixed precision with TF32 enabled; exponential moving average. \\
Training duration
& 3,000 optimizer updates, corresponding to 96,000 prompt groups and
2,304,000 generated images. \\
VVR objective
& Weights $(0.55,0.15,0.15,0.10,0.05)$ for object fidelity,
relations, forbidden content, unmatched components, and background,
respectively, multiplied by the penalty factor $\psi$ in
Eq.~\ref{eq:vvr-training-reward}. \\
VVR complexity and data
& Complexity follows Eq.~\ref{eq:vvr-complexity}. VVR-Easy contains 100,000
unique tasks with $C\leq20$, each drawn from one
generation stratum and containing at most one relation. VVR-Matched contains 100,000 unique tasks matched
to the benchmark distribution over constraint family, family count, and
complexity. \\
\bottomrule
\end{tabular}

\vspace{5pt}
\begin{tabular}{p{0.28\textwidth}p{0.46\textwidth}p{0.18\textwidth}}
\toprule
Training condition & Prompt-group and reward allocation & VVR tasks consumed \\
\midrule
VVR-Easy & 100\% VVR-Easy & 96,000 \\
VVR-Matched & 100\% VVR-Matched & 96,000 \\
GenEval2 & 100\% GenEval2 & 0 \\
GenEval2 $+$ VVR-Easy & 50\% GenEval2, 50\% VVR-Easy & 48,000 \\
GenEval2 $+$ VVR-Matched & 50\% GenEval2, 50\% VVR-Matched & 48,000 \\
OCR & 100\% OCR & 0 \\
OCR $+$ VVR-Easy & 50\% OCR, 50\% VVR-Easy & 48,000 \\
Five-reward & 20\% each: GenEval, GenEval2, PickScore, OCR, UnifiedReward & 0 \\
Five-reward $+$ VVR-Easy & $1/6$ each: the five rewards at left and VVR-Easy & 16,000 \\
\bottomrule
\end{tabular}
\end{table}

%% file: tables/vvr_bench_sd35_posttraining.tex
\begin{table}[H]
\centering
\scriptsize
\setlength{\tabcolsep}{2pt}
\caption{\bench accuracy (\%) of SD3.5 M after post-training.
Training on VVR raises accuracy from 2.81\% to 28.27\% with Easy tasks and 46.60\% with Matched tasks.
Matched training gives the largest gains at high complexity.
Adding VVR-Easy to GenEval2, OCR, or the five-reward objective raises accuracy by a factor of three to seven.
Subscripts are 95\% confidence margins.}
\label{tab:vvr-bench-sd35-posttraining}
\begin{tabular}{lrrrrrr}
\toprule
Training reward & Accuracy (\%) $\uparrow$ & $C_1$ & $C_2$ & $C_3$ & $C_4$ & $C_5$ \\
\midrule
SD3.5-M (pretrained) & 2.81$_{\pm 0.34}$ & 12.01$_{\pm 1.47}$ & 1.19$_{\pm 0.59}$ & 0.35$_{\pm 0.37}$ & 0.05$_{\pm 0.23}$ & 0.00$_{\pm 0.19}$ \\
GenEval2 & 3.87$_{\pm 0.40}$ & 15.42$_{\pm 1.61}$ & 2.39$_{\pm 0.78}$ & 0.91$_{\pm 0.52}$ & 0.05$_{\pm 0.23}$ & 0.05$_{\pm 0.23}$ \\
GenEval2 + VVR-Easy & 21.82$_{\pm 0.82}$ & 54.51$_{\pm 2.15}$ & 32.05$_{\pm 2.12}$ & 14.08$_{\pm 1.60}$ & 6.41$_{\pm 1.16}$ & 1.10$_{\pm 0.56}$ \\
VVR-Easy & 28.27$_{\pm 0.89}$ & \textbf{67.72}$_{\pm 2.04}$ & 45.45$_{\pm 2.23}$ & 17.51$_{\pm 1.73}$ & 8.36$_{\pm 1.29}$ & 1.35$_{\pm 0.60}$ \\
VVR-Matched & \textbf{46.60}$_{\pm 0.98}$ & 67.68$_{\pm 2.04}$ & \textbf{59.17}$_{\pm 2.21}$ & \textbf{45.62}$_{\pm 2.20}$ & \textbf{38.39}$_{\pm 2.15}$ & \textbf{21.82}$_{\pm 1.86}$ \\
GenEval2 + VVR-Matched & 33.50$_{\pm 0.93}$ & 57.93$_{\pm 2.13}$ & 47.27$_{\pm 2.23}$ & 32.34$_{\pm 2.09}$ & 20.32$_{\pm 1.82}$ & 9.22$_{\pm 1.34}$ \\
OCR & 3.58$_{\pm 0.38}$ & 15.37$_{\pm 1.61}$ & 1.51$_{\pm 0.65}$ & 0.40$_{\pm 0.39}$ & 0.05$_{\pm 0.23}$ & 0.00$_{\pm 0.19}$ \\
OCR + VVR-Easy & 24.74$_{\pm 0.86}$ & 57.97$_{\pm 2.13}$ & 36.73$_{\pm 2.18}$ & 18.41$_{\pm 1.76}$ & 7.76$_{\pm 1.26}$ & 1.94$_{\pm 0.70}$ \\
\midrule
Five-reward & 4.83$_{\pm 0.44}$ & 19.16$_{\pm 1.75}$ & 3.12$_{\pm 0.87}$ & 0.91$_{\pm 0.52}$ & 0.30$_{\pm 0.35}$ & 0.00$_{\pm 0.19}$ \\
Five-reward + VVR-Easy & 15.81$_{\pm 0.73}$ & 45.05$_{\pm 2.14}$ & 21.71$_{\pm 1.90}$ & 7.65$_{\pm 1.25}$ & 2.95$_{\pm 0.84}$ & 0.70$_{\pm 0.47}$ \\
\bottomrule
\end{tabular}
\end{table}

%% file: tables/vvr_partial_joint.tex
\begin{table}[H]
\centering
\small
\setlength{\tabcolsep}{4pt}
\caption{Partial and joint constraint satisfaction on \bench by complexity range.
Partial scores are the mean graded count and relation scores; ``all'' is the fraction of tasks in which every count or every relation is satisfied.
Relation columns use only tasks with at least one relation.}
\label{tab:vvr-partial-joint}
\begin{tabular}{@{}llrrrrr@{}}
\toprule
& & \multicolumn{2}{c}{Counts} & \multicolumn{2}{c}{Relations} & \\
\cmidrule(lr){3-4}\cmidrule(lr){5-6}
Range & Model & Partial & All & Partial & All & Accuracy (\%) \\
\midrule
$C_1$ & Pretrained & 0.76 & 0.46 & 0.29 & 0.19 & 12.01 \\
 & VVR-Easy & 0.98 & 0.92 & 0.80 & 0.58 & 67.72 \\
 & VVR-Matched & 0.98 & 0.93 & 0.80 & 0.57 & 67.68 \\
\midrule
$C_2$ & Pretrained & 0.70 & 0.17 & 0.25 & 0.13 & 1.19 \\
 & VVR-Easy & 0.96 & 0.79 & 0.75 & 0.51 & 45.45 \\
 & VVR-Matched & 0.97 & 0.84 & 0.80 & 0.53 & 59.17 \\
\midrule
$C_3$ & Pretrained & 0.69 & 0.10 & 0.20 & 0.06 & 0.35 \\
 & VVR-Easy & 0.94 & 0.66 & 0.59 & 0.27 & 17.51 \\
 & VVR-Matched & 0.98 & 0.88 & 0.77 & 0.41 & 45.62 \\
\midrule
$C_4$ & Pretrained & 0.66 & 0.06 & 0.21 & 0.03 & 0.05 \\
 & VVR-Easy & 0.91 & 0.44 & 0.61 & 0.22 & 8.36 \\
 & VVR-Matched & 0.98 & 0.76 & 0.82 & 0.42 & 38.39 \\
\midrule
$C_5$ & Pretrained & 0.61 & 0.02 & 0.21 & 0.02 & 0.00 \\
 & VVR-Easy & 0.87 & 0.15 & 0.57 & 0.13 & 1.35 \\
 & VVR-Matched & 0.97 & 0.48 & 0.85 & 0.41 & 21.82 \\
\bottomrule
\end{tabular}
\end{table}

%% file: tables/sd35_postfreeze_external_full.tex
\begin{table}[H]
\centering
\scriptsize
\setlength{\tabcolsep}{2.7pt}
\caption{Complete final evaluation of the pretrained model and nine post-training conditions.
PickScore uses the Pick-a-Pic v1 validation benchmark and HPSv2.1 uses the official four-domain HPDv2 benchmark.
HPSv3, CLIPScore, aesthetic score, ImageReward, and UnifiedReward are macro-averaged across the shared cross-domain panel.
VVR accuracy, GenEval, GenEval2, and OCR retain their native units.
Every HPSv3 value, including the pretrained one, is the mean over the four cross-domain prompt sets.
Images of the pretrained model use sampling seed 42; images of trained models use seed 20260912, except for PickScore and HPSv2.1, which use seed 42 for every model.}
\label{tab:sd35-postfreeze-external-full}
\resizebox{\textwidth}{!}{%
\begin{tabular}{lrrrrrrrrrrr}
\toprule
Training reward & VVR & GenEval & GenEval2 & OCR & PickScore & HPSv2.1 & HPSv3 & CLIPScore & Aesthetic & ImageReward & UnifiedReward \\
\midrule
Pretrained & 0.028 & 0.616 & 0.237 & 0.476 & 0.841 & 0.300 & 7.689 & 0.956 & 5.517 & 0.929 & 0.636 \\
\midrule
GenEval2 & 0.039 & 0.688 & 0.454 & 0.501 & 0.848 & 0.297 & 8.289 & 0.974 & 5.523 & 1.110 & 0.635 \\
GenEval2 + VVR-Easy & 0.218 & 0.718 & 0.478 & 0.532 & 0.848 & 0.302 & 8.511 & 0.976 & 5.533 & 1.155 & 0.637 \\
VVR-Easy & 0.283 & 0.729 & 0.268 & 0.587 & 0.849 & 0.294 & 8.275 & 0.979 & 5.483 & 1.114 & 0.641 \\
$\Delta$ add VVR-Easy & +0.180 & +0.030 & +0.025 & +0.030 & +0.0004 & +0.005 & +0.222 & +0.002 & +0.009 & +0.045 & +0.002 \\
\midrule
VVR-Matched & 0.466 & 0.709 & 0.360 & 0.536 & 0.847 & 0.291 & 7.997 & 0.980 & 5.472 & 1.109 & 0.636 \\
GenEval2 + VVR-Matched & 0.335 & 0.712 & 0.491 & 0.510 & 0.848 & 0.301 & 8.498 & 0.972 & 5.516 & 1.148 & 0.637 \\
$\Delta$ add VVR-Matched & +0.296 & +0.024 & +0.038 & +0.009 & $-0.0002$ & +0.004 & +0.209 & $-0.001$ & $-0.007$ & +0.038 & +0.001 \\
\midrule
OCR & 0.036 & 0.625 & 0.225 & 0.962 & 0.844 & 0.290 & 7.601 & 0.960 & 5.477 & 0.994 & 0.633 \\
OCR + VVR-Easy & 0.247 & 0.678 & 0.252 & 0.941 & 0.846 & 0.290 & 7.765 & 0.969 & 5.488 & 1.087 & 0.636 \\
$\Delta$ add VVR-Easy & +0.212 & +0.053 & +0.027 & $-0.021$ & +0.002 & 0.000 & +0.164 & +0.009 & +0.012 & +0.092 & +0.002 \\
\midrule
Five-reward & 0.048 & 0.739 & 0.342 & 0.856 & 0.850 & 0.294 & 8.137 & 0.974 & 5.510 & 1.125 & 0.640 \\
Five-reward + VVR-Easy & 0.158 & 0.751 & 0.383 & 0.826 & 0.848 & 0.300 & 8.444 & 0.976 & 5.541 & 1.152 & 0.640 \\
$\Delta$ add VVR-Easy & +0.110 & +0.012 & +0.041 & $-0.030$ & $-0.002$ & +0.006 & +0.307 & +0.002 & +0.031 & +0.027 & +0.0003 \\
\bottomrule
\end{tabular}%
}
\end{table}

%% file: tables/sd35_external_bootstrap.tex
\begin{table}[H]
\centering
\scriptsize
\setlength{\tabcolsep}{4pt}
\caption{Effect of adding VVR to an existing reward, with paired bootstrap 95\% intervals over prompts (10,000 resamples; stratified by prompt set for macro-averaged metrics and by domain for HPSv2.1).
The GenEval, GenEval2, OCR, HPSv3, and UnifiedReward evaluations retained only aggregate scores, so they have no intervals.}
\label{tab:external-bootstrap}
\resizebox{\linewidth}{!}{%
\begin{tabular}{lccccc}
\toprule
Contrast & PickScore & HPSv2.1 & CLIPScore & Aesthetic & ImageReward \\
\midrule
GenEval2 $+$ VVR-Easy $-$ GenEval2 & $+0.0004$ {\tiny[$-0.0012$, $0.0021$]} & $+0.0047$ {\tiny[$0.0042$, $0.0052$]} & $+0.0023$ {\tiny[$-0.0014$, $0.0058$]} & $+0.009$ {\tiny[$-0.003$, $0.022$]} & $+0.045$ {\tiny[$0.025$, $0.066$]} \\
GenEval2 $+$ VVR-Matched $-$ GenEval2 & $-0.0002$ {\tiny[$-0.0019$, $0.0015$]} & $+0.0039$ {\tiny[$0.0034$, $0.0045$]} & $-0.0011$ {\tiny[$-0.0044$, $0.0021$]} & $-0.007$ {\tiny[$-0.020$, $0.006$]} & $+0.038$ {\tiny[$0.020$, $0.057$]} \\
OCR $+$ VVR-Easy $-$ OCR & $+0.0018$ {\tiny[$-0.0002$, $0.0037$]} & $0.0000$ {\tiny[$-0.0007$, $0.0006$]} & $+0.0086$ {\tiny[$0.0054$, $0.0118$]} & $+0.012$ {\tiny[$-0.001$, $0.025$]} & $+0.092$ {\tiny[$0.071$, $0.113$]} \\
Five-reward $+$ VVR-Easy $-$ Five-reward & $-0.0024$ {\tiny[$-0.0041$, $-0.0006$]} & $+0.0060$ {\tiny[$0.0054$, $0.0066$]} & $+0.0023$ {\tiny[$-0.0009$, $0.0055$]} & $+0.031$ {\tiny[$0.020$, $0.042$]} & $+0.027$ {\tiny[$0.009$, $0.046$]} \\
\bottomrule
\end{tabular}
}
\end{table}

%% file: tables/human_annotator_agreement.tex
\begin{table}[H]
\centering
\small
\caption{Agreement between annotator pairs on the 400 preference pairs.
Labels are decoded to the preferred model before comparison.
``Both chose'' uses only the pairs on which neither annotator chose a tie; ``ties as a label'' uses all 400 pairs with tie as a third label, which is also the label set of Cohen's $\kappa$.
Fleiss' $\kappa$ over the three annotators is 0.49.}
\label{tab:human-annotator-agreement}
\begin{tabular}{@{}lccc@{}}
\toprule
Annotator pair & Agreement, both chose (\%) & Agreement, ties as a label (\%) & Cohen's $\kappa$ \\
\midrule
1 vs.\ 2 & 84.9 {\scriptsize(298/351)} & 76.0 {\scriptsize(304/400)} & 0.50 \\
1 vs.\ 3 & 81.9 {\scriptsize(276/337)} & 72.0 {\scriptsize(288/400)} & 0.44 \\
2 vs.\ 3 & 84.6 {\scriptsize(312/369)} & 78.8 {\scriptsize(315/400)} & 0.53 \\
\midrule
Mean & 83.8 & 75.6 & 0.49 \\
\bottomrule
\end{tabular}
\end{table}